\documentclass[acmtog, nonacm]{acmart}

\AtBeginDocument{%
  \providecommand\BibTeX{{%
    \normalfont B\kern-0.5em{\scshape i\kern-0.25em b}\kern-0.8em\TeX}}}

\acmJournal{TOG}

\usepackage{booktabs}
\usepackage{savesym}
\savesymbol{zifour@default}
\savesymbol{zifour@scaled}
\usepackage{enumerate}
\usepackage{multirow} 
\usepackage[normalem]{ulem}
\usepackage{inconsolata}

\usepackage[ruled]{algorithm2e}

\usepackage{listings}
\usepackage{xcolor}
\definecolor{codebg}{RGB}{230,212,194} 
\SetAlFnt{\small}
\SetAlCapFnt{\small}
\SetAlCapNameFnt{\small}
\SetAlCapHSkip{0pt}

\usepackage{xspace}
\usepackage{cleveref}

\newcommand{\denselist}{\itemsep 0pt\parsep=0pt\partopsep 0pt\vspace{-\topsep}}

\newcommand{\methodname}{\text{ShapeLib}\xspace}

\newcommand{\llmbaseline}{\text{LLM-Only}\xspace}

\begin{document}

\title[\methodname: Designing a library of programmatic 3D shape abstractions with Large Language Models]{\methodname: Designing a library of programmatic 3D shape abstractions \\with Large Language Models}

\author{R. Kenny Jones}
\email{rkjones4@stanford.edu}
\affiliation{%
    \institution{Stanford University}
    \country{USA}
}

\author{Paul Guerrero}
\email{guerrero@adobe.com}
\affiliation{%
    \institution{Adobe Research}
    \country{United Kingdom}
}

\author{Niloy J. Mitra}
\email{n.mitra@cs.ucl.ac.uk}
\affiliation{%
    \institution{University College London and Adobe Research}
    \country{United Kingdom}
}

\author{Daniel Ritchie}
\email{daniel\_ritchie@brown.edu}
\affiliation{%
    \institution{Brown University}
    \country{USA}
}

\begin{abstract}

We present \textit{\methodname}, the first method that uses the priors of Large Language Models (LLMs) to design libraries of programmatic 3D shape abstractions.
Our system accepts two forms of user-provided design intent: high-level text descriptions of functions to include in the output library and a small seed set of exemplar shapes.
We discover a library of abstractions that matches this design intent with a guided LLM workflow that first \textit{proposes} different ways of applying and implementing functions, and then \textit{validates} these functions are helpful in representing seed set shapes. 
To extend beyond the seed set, we develop library-specific recognition networks that map shapes (represented as primitives, voxels, or point clouds) to programs that use these newly discovered abstractions.
Across multiple modeling domains (split by shape category), we find that LLMs, when thoughtfully combined with geometric reasoning, can be guided to author libraries of abstraction functions 
that generalize across shape distributions.
Our framework takes a step towards realizing the long-standing shape analysis aspiration of discovering reusable, programmatic shape abstractions while exposing interpretable, semantically aligned interfaces.
Our extensive evaluation demonstrates that \methodname provides distinct advantages over prior alternative abstraction discovery works in terms of generalization, usability, and maintaining plausibility under manipulation. 
Finally, we demonstrate that \methodname's abstraction functions unlock a number of downstream applications, combining LLM reasoning over shape programs with geometry processing tools to support shape editing and generation workflows.

\end{abstract}


\begin{teaserfigure}
\centering
  \includegraphics[width=\linewidth]{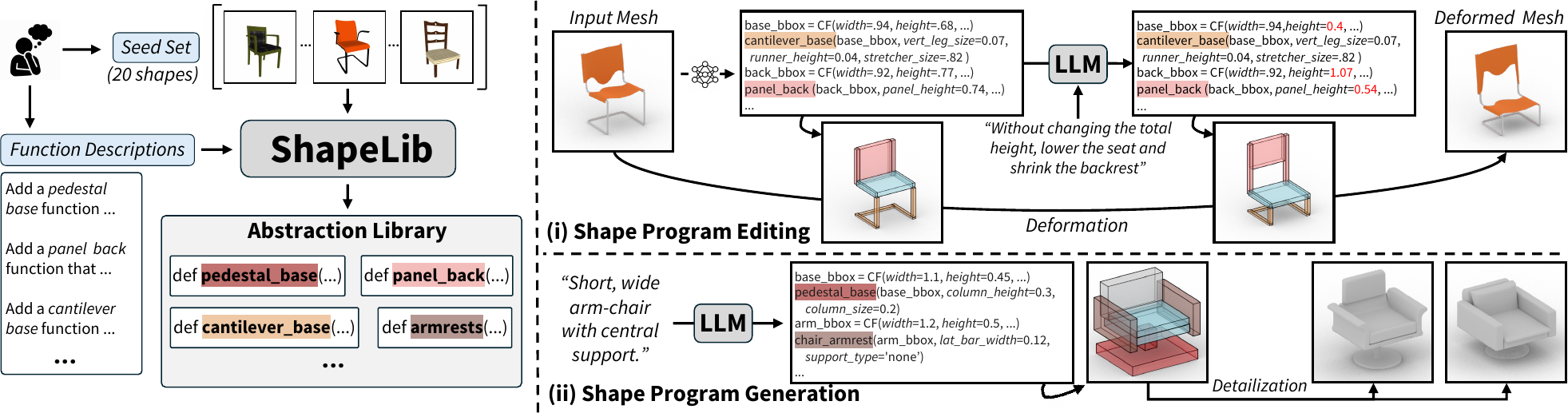}
  \caption{
  \methodname guides an LLM to design a library of programmatic shape abstraction functions from a small set of seed shapes and function descriptions.
  These abstraction functions generalize to new shapes and expose semantically aligned, easy-to-work with interfaces, supporting downstream tasks like (i) shape editing (through LLM program editing and mesh deformation) and (ii) shape generation (through LLM program synthesis and detailization).
  }
  \label{fig:teaser}
\end{teaserfigure}

\maketitle

\section{Introduction}

3D shapes are central to many visual computing problems, where applications 
depend on the ability to edit, manipulate, analyze, and synthesize 3D assets. 
Procedural models---structured programs that produce geometry when executed---are an appealing representation for 3D shapes that provide natural support for these operations. 
Well-designed procedural models expose (semantic) handles that end-users can interpret and use to easily manipulate output geometry. 
Good programs, however, are expensive to author or acquire.

\begin{figure}[t!]
\centering
 \includegraphics[width=\linewidth]{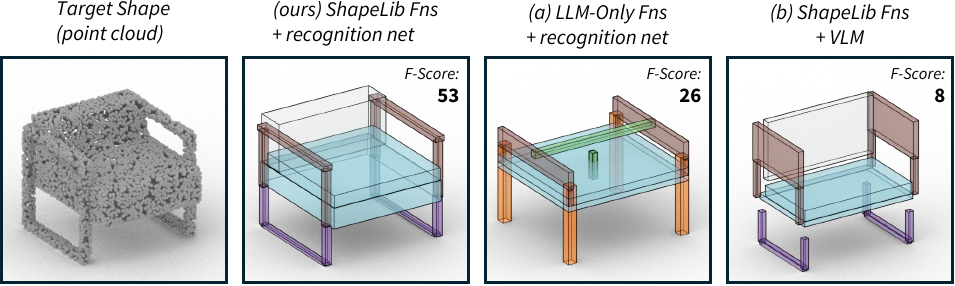}
\caption{
Motivation. 
\methodname combines LLM reasoning with geometric validation to discover abstractions that can be applied to reconstruct new shapes with library-specific recognition networks. However, when guidance is lacking, current LLMs still struggle to produce effective programmatic shape representations. Naively relying on LLMs to either (a) author library functions without validating against a seed set or (b) accurately apply abstractions to new shape instances, produces significantly worse results.
} 
\label{fig:llm_motiv}
\end{figure}

When experts design procedural models, they rely on libraries of modeling functions that expose the right level of abstraction for a particular domain.
For instance, authoring a good procedural model of a building might require access to programmatic abstractions that tile windows over a facade~\cite{wonka2003instant}, or automatically extrude common types of roofing patterns from a boundary~\cite{muller2006procedural}.
When interpretable modeling functions are not available, this shape modeling task becomes more difficult (see Figure~\ref{fig:llm_motiv}).

\begin{figure*}[t!]
\centering
\includegraphics[width=\linewidth]{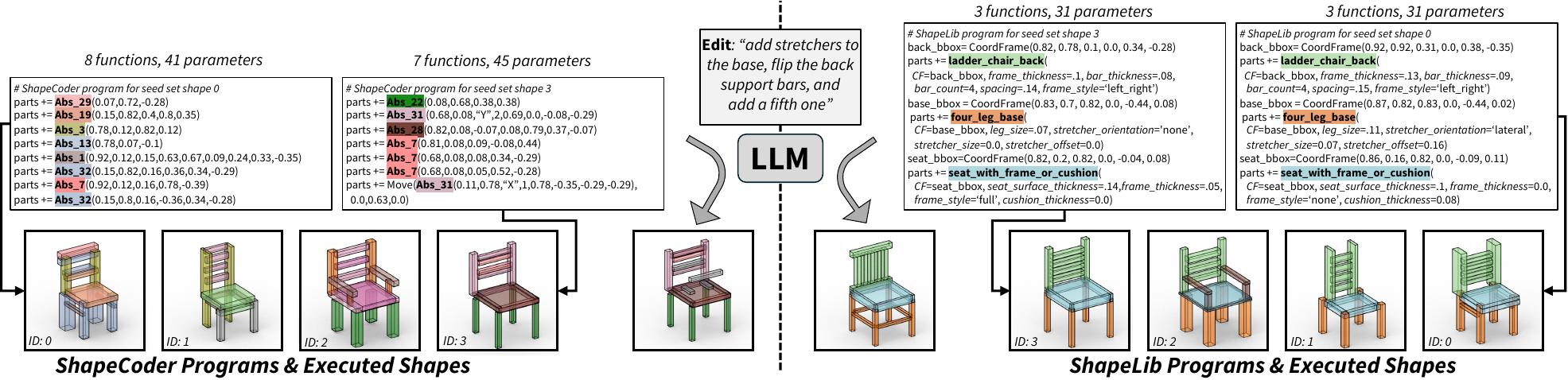}
\caption{
Semantic alignment. 
We illustrate the benefits of using semantic guidance in library learning methods by running two different methods on the same seed shapes and comparing the programs found by these methods to represent some of these seed shapes (see the \textit{ID} annotation for each reconstruction). 
ShapeCoder~\cite{jones2023shapecoder} (\textit{left}) discovers abstractions without semantic guidance; making downstream tasks like shape editing difficult (\textit{middle-left}). 
\methodname (\textit{right}) exposes an interpretable interface, with semantically named parameters, and applies functions in a consistent fashion; simplifying tasks like shape editing (\textit{middle-right}).  
} 
\label{fig:prog_comp}
\end{figure*}

Unfortunately, designing a library of procedural modeling functions is significantly more challenging than authoring a single procedural model.
Despite this difficulty, researchers have investigated how to automatically discover libraries of programmatic abstractions~\cite{jones2023shapecoder,ellis2021dreamcoder}  with techniques that are unaware of shape semantics.
Starting with a dataset of shapes and a base modeling language with elementary functions, these bottom-up methods greedily grow libraries by defining new functions based on how well they compress patterns over the dataset.
While these methods successfully optimize their compression-based objectives, they develop their libraries without any semantic guidance, so the functions they produce only align to shape semantics by chance, 
making it difficult for users to understand or manipulate the resulting programmatic interface (see Figure~\ref{fig:prog_comp}).

As an alternative, we investigate how Large Language Models (LLMs) can help with this programmatic abstraction design problem.
LLMs have demonstrated remarkable success over a surprisingly diverse range of tasks, from 3D layout synthesis~\cite{hu2024scenecraft,littlefair2025flairgpt} to general code generation~\cite{jiang2024survey}.
There are reasons to believe they might be useful in helping to design programmatic shape abstractions: they have world knowledge about the semantic relationships of parts within shapes and are proficient at writing code.
At the same time, LLMs also have limitations that temper their usefulness for reasoning over programmatic shape representations (see Figure~\ref{fig:llm_motiv}).
As we demonstrate experimentally, LLMs still struggle to understand complex geometric layouts and often misinterpret or misattribute constraints and relations between parametric controls.

To leverage LLMs, while avoiding these potential pitfalls, we propose \methodname, a hybrid system that guides an LLM through the creation of a library of programmatic shape abstraction functions from a specified design intent.
For a given shape modeling domain (e.g., a shape category), a user provides input to \methodname for how the abstraction library should be designed.
We elicit this design intent in two forms: (i) high-level function descriptions in natural language, and (ii) a small seed set of exemplar shapes.
The two modalities are complementary: the first mode allows the user to specify the kinds of functions they would like to interface with; while the second mode provides geometric references that guide and constrain library development.
\methodname discovers functions that produce cuboid primitives that represent semantically aligned part bounding boxes. 
Instead of trying to completely reproduce surface geometry, we generate structured shape representations that support multiple downstream tasks (see \Cref{fig:teaser}). 

\methodname breaks the complex library design process into a series of sub-problems. 
First, we use an LLM to convert the user-provided function descriptions into a semantically aligned structured library interface.
Next, we task an LLM with proposing a set of possible applications of these functions to explain shapes from the seed set.
We then use these proposed applications to automatically formulate input/output examples that guide an LLM, in turn, to propose a set of possible function implementations.
We finalize the library with a validation step that performs a geometric analysis over the proposed sets of possible function implementations and their applications.
This pipeline ensures that the outputs of LLM-produced abstraction functions that get added into the library are grounded by patterns observed in the seed set.  

To decide how these functions should be used to represent shapes outside of the seed set, we train \textit{recognition networks} that learn to map input shapes to output programs that use the abstraction functions.
To train this network, we task the LLM with creating a synthetic data generation procedure that uses the library functions to sample random programs that can be executed to form approximately correct part layouts.
In this way, even starting from a small seed set, \methodname can author programs that use these abstraction functions to explain a much larger collection of shapes. 

We evaluate \methodname by using it to design libraries of programmatic abstraction functions over multiple shape modeling domains, which we split by category (\texttt{chair}, \texttt{table}, \texttt{storage}, \texttt{lamp}, \texttt{faucet}).
Compared with alternative methods that discover libraries of shape abstractions, we experimentally demonstrate the benefits of \methodname over a number of axes, including:
(i)~\textit{Generalization}: abstractions are useful for modeling shapes outside of the seed set; (ii)~\textit{Usability}: abstractions expose an interpretable interface that is well-aligned with semantics, and hence easy to use; and  (iii)~\textit{Plausibility}: abstractions constrain outputs to maintain shape semantics under manipulation.

\methodname relies on a novel blend of LLM guidance and geometric reasoning to outperform existing alternatives.
Without semantic guidance, state-of-the-art geometry based systems like ShapeCoder \cite{jones2023shapecoder} find abstractions that can generalize to new shapes, but are hard to interact with and produce non-semantic outputs under parameter modifications. 
Without a \textit{seed set} of reference geometries, 
LLMs can convert \textit{function descriptions} into seemingly sensible abstraction functions, 
but their implementations have structural inconsistencies which limits their ability to represent actual shape structures.
In contrast, \methodname exploits the two complementary design intent modalities, guiding an LLM through the process of authoring programmatic shape abstractions.

We find that the abstraction libraries produced by \methodname support a range of applications, as depicted in Figure~\ref{fig:teaser}.
First, we can use our recognition networks to convert unstructured input geometry into well-structured shape programs.
Second, we demonstrate that \methodname's functions facilitate shape editing: an LLM can modify an inferred program, and the executed geometry of the edited program can be used to deform the original shape. 
Finally, we show that our function library can aid in text-to-shape generation pipelines: we use an LLM to convert a text prompt into a shape program, and then we convert this structured representation into detailed geometry.

In summary, our contributions are:
\begin{enumerate}[(i)]
    \denselist
    \item \methodname, the first approach that, 
    given a seed set and textual function descriptions as input, guides an LLM through the development of a library of programmatic shape abstraction functions that are reusable and semantically aligned. 
    \item Recognition networks that learn from LLM-authored synthetic data samplers to infer programs using the discovered functions.  
    \item Demonstrations of how programmatic shape abstractions support applications like shape editing and generation.
\end{enumerate}

\section{Related Work}

\textit{\textbf{Library Learning. }}
The goal of library learning is to automatically discover a set of useful programmatic abstractions. 
Several prior works~\cite{ellis2021dreamcoder, ellis2018library, babble} have studied this task for general program synthesis domains. 
These methods take as input a set of simple tasks or programs that use only basic operators and find a library of more abstract functions that can represent the inputs more compactly.
Though these approaches have demonstrated impressive generality, their non-specialization has limited their usability for more complex 3D modeling domains. 
ShapeMOD~\cite{jones2021shapeMOD} and ShapeCoder~\cite{jones2023shapecoder} extend this library learning machinery to 3D shapes, using specialized search strategies over parametric expressions.
However, unlike \methodname's use of an LLM prior, all of the above methods derive the library based \textit{only} on the input examples. 
As we find experimentally, this limits the \emph{interpretability} of the functions discovered by these methods, which produce interfaces that are hard to work with and align poorly with semantics.
Lilo~\cite{grand2024lilo} is a recent approach that also leverages an LLM prior for general library learning. 
However, it only uses the LLM to \emph{name} functions found by a non-semantic compression-based method, so the prior issues still remain.

\noindent\textit{\textbf{Shape Program Synthesis. }}
A number of learning-based methods have studied the shape program synthesis problem for specific low-level languages, such as 
CSG modeling~\cite{ren2021csg, kania2020ucsg}, 
CAD workflows~\cite{Fusion360Gallery,wu2021deepcad},
L-systems~\cite{guo2020inverse, lee2023latent}.
Other approaches have investigated how to extend this machinery to work over more general visual programming domains~\cite{sharma2018csgnet,jones2024VPIEdit,ganeshan2023improving}. While these methods assume that the domain-specific languages are fixed and known ahead of time, \methodname aims to discover a library of new abstraction functions directly.

The visual programs we consider represent complex shapes through a combination of simple part primitives; many inverse modeling works have taken a similar framing~\cite{jones2020shapeAssembly,PlankAssembly,StructureNet,GRASS}.
Unlike these prior works, \methodname leverages an LLM to create compact, semantic programmatic interfaces for these structured representations.
Prior work has found that shape programs that use only general, low-level operations hurt performance on downstream tasks and expose interfaces that users find more difficult to manipulate~\cite{jones2021shapeMOD}.
A possible solution is first to have an expert user author complex procedural models capable of representing large shape families and then to train networks that learn how to parameterize these programs ~\cite{pearl2022geocode,infinigen2023infinite,infinigen2024indoors}.
This approach can produce compelling results, but it is expensive and time-consuming to ask an expert to craft well-designed procedural functions.
In contrast, \methodname explores how an LLM can be guided through the process of automatically designing useful abstraction functions from easily provided, high-level design intent.

\noindent\textit{\textbf{LLMs for 3D Generation and Editing.}}
LLMs have recently been used to directly generate scene layouts from text prompts~\cite{yang2024holodeck, feng2023layoutgpt, zhang2024scene, littlefair2025flairgpt, aguina2024open, tam2024scenemotifcoder}.
These methods take an opposite approach to those in the prior paragraph: they rely almost wholly on the LLM prior, and only extract limited information from the seed set (e.g., as in-context examples).
Though similar at surface level, shape modeling and scene modeling present different challenges, as part-to-part relations in shapes have more constrained and specific relationships than object-to-object relations in most scenes. 
We find that ablated versions of \methodname that minimize reliance on the prior provided by seed set perform significantly worse.

Relatedly, some recent work has explored how LLMs might enable visual editing workflows.
ParSEL~\cite{ganeshan2024parsel} uses an LLM to convert a structured shape representation into an approximate procedural model that supports a particular user-specified edit.
BlenderAlchemy~\cite{huang2024blenderalchemy} uses an LLM to modify existing procedural shapes and scenes encoded as blender programs.
3D-GPT~\cite{sun20233d} uses an expert-designed library of procedural modeling functions~\cite{infinigen2023infinite} to synthesize and edit virtual worlds. 

While the results of these methods are exciting, their LLM integration is limited by their reliance on highly structured inputs, such as hand-crafted procedural models that use expert-designed functions.
In contrast, \methodname operates over more flexible inputs, and aims to automatically discover new programmatic shape abstractions that can generalize to explain new shapes.

\begin{figure*}[t!]
\centering
  \includegraphics[width=\linewidth]{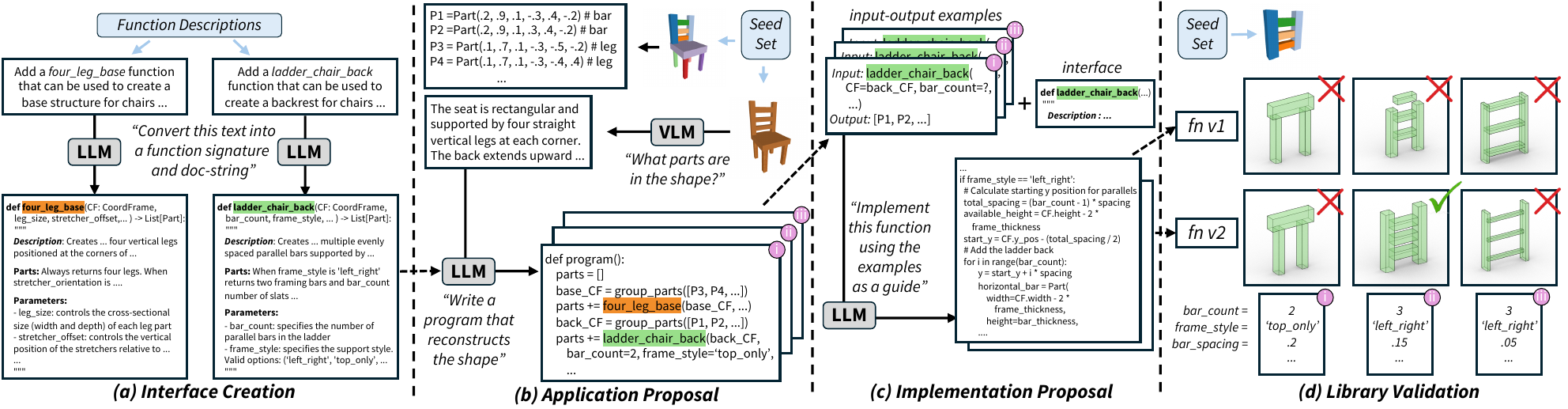}
   \caption{\methodname overview. We design a function library in four steps, starting from function descriptions and a set of seed shapes. First, (a) we prompt an LLM to create function interfaces that define parameters and annotate the function's purpose. Then, (b) the LLM is prompted to propose possible applications of the functions that reconstruct the seed set shapes. Next, (c) we use this information to guide the LLM to propose possible function implementations. The library is finalized with a validation step (d) that searches for pairs of  applications and implementations that best reconstruct the seed set shapes.} 
  \label{fig:method_fig}
\end{figure*}

Finally, there has been a growing interest in \textit{finetuning} LLMs to produce shape programs, especially for CAD domains. 
Such approaches have explored unified multimodal conditioning~\cite{CADMLLM}, synthesizing editable models from unconstrained images~\cite{CADCrafter}, or improving LLM performance through self-improvement mechanisms~\cite{BlenderLLM}.
In \methodname, we assume that the LLM is frozen, and design a method that guides the model through producing a library of grounded abstraction functions, rather than only creating programs for individual shapes.

\section{Overview}
\label{sec:overview}

\methodname guides an LLM through the process of developing a library of programmatic abstraction functions from an input design intent.
In our framing, we assume that a user has a shape modeling domain in mind (e.g., a shape category), and communicates this design intent to our system with two modalities: function descriptions and a seed set of shapes. 
Each function description specifies the intended behavior of one abstraction function that should be reusable across a distribution of shapes.
Correspondingly, the seed set is meant to provide representative geometric grounding for that broader distribution.
Our goal is not to require the discovered library to exhaustively capture every structural variant present in those examples, but rather to cover a broad and useful range of variation within the category.
We provide illustrative examples of this design intent in Appendix~\ref{sec:app_design_intent}.

Each seed set we consider is composed of 20 shapes with part-level semantic segmentations and textured renders.
\methodname finds programmatic representations of these shape structures, with functions that output part bounding proxies.
In this way, our programmatic abstractions also provide a layer of geometric abstraction.

\methodname's framing receives a number of benefits from the prior knowledge encoded in LLMs.
Through extensive pretraining on large-scale code repositories and targeted finetuning on diverse code-related tasks, these models have demonstrated exceptional proficiency in complex program synthesis settings~\cite{openai2025competitiveprogramminglargereasoning}.
Moreover, because they are trained on human-authored code, they can generate functions with meaningful names and parameters, resulting in interfaces that are both interpretable and easy to use.
However, for tasks requiring complex spatial reasoning, LLMs without explicit grounding remain prone to hallucination, often producing implausible layouts or distributions that diverge from realistic geometric structures.

The different design intent modalities provide complementary signals that work together to help ground \methodname.
Describing desired function properties allows a user to exercise control over the interface exposed by the library's abstractions.
Moreover, it also provides guidance for the LLM, attuning the model towards a particular way of solving the shape abstraction task.
Whereas the function description allows the user to provide guidance through \textit{language}, the seed set allows the user to provide \textit{geometric} guidance.
This not only offers an additional control mechanism for design considerations that are hard to express textually, it also provides an avenue for validation; 
\methodname evaluates the plausibility of its productions by searching for function implementations and applications that can explain sub-structures in the seed set shapes.
This validation step assumes that these design intents complement one another -- for every function description, there should be some shapes in the seed set that exemplify the corresponding pattern.

In Sec.~\ref{sec:lib_design}, we describe how we convert user-provided design intent into a fully realized library of abstraction functions.
In Sec.~\ref{sec:lib_usage}, we describe how we can expand the usage of this library beyond the seed set by training a recognition network on synthetic data.

\section{Designing a Library of Shape Abstractions}
\label{sec:lib_design}

We depict \methodname's pipeline for converting a user's input design intent into a library of abstraction functions in Figure~\ref{fig:method_fig}.
This process is divided into four stages:
 (a)~\textit{interface creation} converts function descriptions into a library interface (Sec.~\ref{sec:lib_interface}); 
(b)~\textit{application proposal} identifies which library functions should model which seed set shapes (Sec.~\ref{sec:prop_apps});  
 (c)~\textit{implementation proposal} generates candidate function implementations (Sec.~\ref{sec:prop_impls});
 (d)~\textit{library validation} finalizes the library through geometric analysis (Sec.~\ref{sec:lib_validation}).

\subsection{Creating the Library Interface}
\label{sec:lib_interface}

\methodname~first converts user function descriptions into a library interface (Fig.~\ref{fig:method_fig}, a).
We prompt an LLM to produce a structured interface, where for each function it produces the function's type signature and an accompanying doc-string.
We provide the LLM with two default classes: a \textit{Part} class defines axis-aligned cuboid primitives that abstract detailed geometry and a \textit{CoordFrame} class defines a local bounding volume.
The LLM produces function signatures that expose parametric handles, e.g., the numbers of bars in a ladder back or the height of a base runner.
Each function is instructed to take in a special first parameter, \textit{CF}, a \textit{CoordFrame} that specifies the expected extents of the function's output. 
Functions are typed so that they return a list of \textit{Part} objects.

Through in-context examples and instructions, we prompt the LLM designed doc-string to have a particular structure. 
First, it defines a \textit{description} field to explain the high-level goals of the function.
Then, it defines a \textit{parts} field, that specifies what kinds and how many parts should be produced depending on the input parameters.
Finally, it defines a \textit{parameter} field, that explains how each variable should affect the output structure.
This interface is then used in subsequent stages to guide the library development.

\subsection{Proposing Function Application }
\label{sec:prop_apps}

As LLMs are prone to hallucinations, we do not directly implement functions following the prior step. 
Instead, we would like to be able to ground each function implementation by making sure it is capable of reproducing some structures from the seed set.
To this end, we first ask the LLM to propose programs that apply the library functions to explain the exemplar shapes (Fig.~\ref{fig:method_fig}, b).

We begin by sampling a shape from the seed set and ask an LLM with vision capabilities (i.e., a VLM) to describe the parts that it sees when given a rendering of the shape.
We also convert the 3D semantic part annotations into a list of labeled \textit{Part} objects.
We combine these inputs together, and task an LLM with writing a program that uses the library functions to recreate the list of Parts. 
As we have not created function implementations yet, the LLM chooses functions and their parameters based only on the interfaces defined previously. 
We ask the LLM to use a special \textit{group\_parts} function when constructing this program, that consumes a list of input \textit{Part} objects and returns a bounding \textit{CoordFrame} object.
With this formulation, we can automatically parse input-output examples; i.e., input parameters and the corresponding parts the functions should generate.

As we later demonstrate empirically, the accuracy of individual LLM calls has a high variance, which makes them hard to trust. 
Therefore, instead of finding a single program for each shape, we run this procedure $K_A$ times for each shape in the seed set ($K_A$=5).

\begin{figure*}[t!]
\centering
  \includegraphics[width=\linewidth]{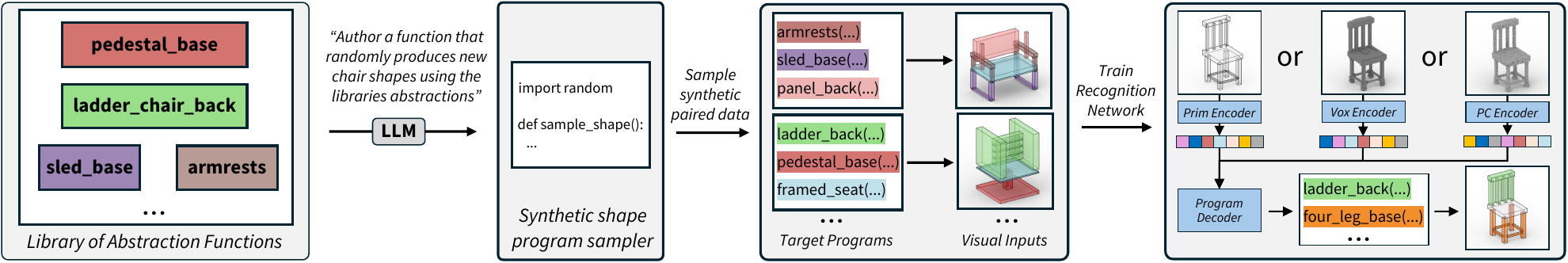}
   \caption{
   Recognition network overview. 
   To extend \methodname's abstractions beyond the shapes from the seed set, we guide the LLM to author a `sample\_shape' function.
   This procedure uses the library functions to sample synthetic shape programs.
   We can execute these programs to get paired data for training our recognition networks that learn how to map shapes (represented as primitives, voxels, or point clouds) to shape programs that use library functions.} 
  \label{fig:net_fig}
\end{figure*}

\subsection{Proposing Function Implementations}
\label{sec:prop_impls}

\methodname~ now has the information it needs to author good function implementations: typed signatures, doc-string guidance, and input-output examples.
From this input, we ask the LLM to complete the implementation of each function so that it matches the signature type, meets the doc-string specification, and respects the observed patterns present in the usage examples (Fig.~\ref{fig:method_fig}, c). 
Of note, we find that the LLM predictions in the application proposal step do a good job of identifying which functions should explain which parts, but are less proficient at predicting reasonable parameter values. 
With this in mind, we mask out parameter values with a special token (`?') when formulating input-output examples for the implementation prompt.
We do this for every parameter value, except for \textit{CF}, as the correct input value for this parameter, with respect to a specific set of outputs \textit{Parts}, can be found automatically with the \textit{group\_parts} function.

Similar to before, we find that implementations authored by the LLM display substantial variance in terms of how well they match the input specification.
So, for each function in our library, we propose $K_I$ different ways that it could be implemented ($K_I$=4).

\subsection{Validating the Library}
\label{sec:lib_validation}

From the prior steps we have (a) function doc-strings and signatures, (b) input-output usage examples, and (c) proposals of how the function should be implemented.
The validation step is responsible for deciding which of these proposals are \textit{grounded} 
and not just LLM hallucinations (Fig~\ref{fig:method_fig}, d).

This process must be able to identify LLM proposals that aren’t able to \emph{well-reconstruct} patterns from the seed shapes, either due to bugs or semantic errors. 
Our notion of \emph{well-reconstruct} uses an error metric that operates over two sets of primitives, comparing corner-to-corner distances (see Section~\ref{sec:app_prim_error}).
Instead of requiring exact geometric matches, we allow some leniency by setting a maximum error threshold, and considering all matches that achieve an error under this threshold as \emph{valid}.

To begin the validation step, for each function, 
we gather all of the LLM generated input-output examples,
and separate them into
a list of input parameter sets, 
and a list of output part sets. 
We use the former list to come up with a much larger set of ways that the function could be parameterized.
Then, we execute the function on all parameter combinations from this set, and for each input set, we check if the function's execution achieves a \emph{valid} match against any of the output part sets under our error metric.
Compared with the alternative of only checking the paired LLM input-output predictions, we observed this simple search strategy was able to successfully recover from LLM mistakes in this difficult reconstruction task.  

At this point, for each group of parts from (b) we record which implementation from (c) best matches the observed part structure.
We keep the implementation that achieves the \textit{best} error across the \textit{most} part groups, and remove all others proposals.
If this \textit{best} implementation found valid applications across multiple seed set shapes, we update the library interface entry with its implementation logic. 
Otherwise, we remove the function entry from the interface.

\section{Library Specific Recognition Networks}
\label{sec:lib_usage}

In Section~\ref{sec:lib_design}, we constructed a library of shape abstraction functions that capture patterns observed in the seed set, but a question remains: how can we use the discovered functions to represent new shapes?
In this section, we describe our strategy for expanding library function usage beyond the seed set with a recognition network, depicted in Figure ~\ref{fig:net_fig}.
This network learns how to solve an inverse task: given an input shape, it reconstructs the shape by writing a program that uses the abstraction functions.
We train our recognition network in a supervised fashion, using paired synthetic data generated by an LLM authored procedure. 
We first describe the generation of this synthetic data sampler, and then describe the design of our recognition network.

\subsection{Generating a Synthetic Shape Program Sampler}
In this step, we once again make use of the strong prior of LLMs by asking it to design a procedure that uses the abstraction functions to randomly synthesize synthetic shape programs.
To accomplish this, we design a prompt that describes the library we have developed, including the interface of each function and examples of how to use the abstractions (sourced from the validation stage).
This prompt asks the LLM to design a \textit{sample\_shape} function that randomly produces new shape programs using the provided abstractions.
Interestingly, we find that frontier LLMs are able to provide useful implementations of such a \textit{sample\_shape} function.
As shown in Figure~\ref{fig:net_fig}, some of these random outputs produce good shape abstractions, while other random samples violate class semantics.
With this in mind, instead of attempting to get the LLM to perfect its implementation, we treat its output as a synthetic data generator for our recognition network. 
To broaden the coverage and variety of structures that these \textit{sample\_shape} functions produce, we employ an iterative refinement loop that provides automatic feedback to the LLM.
This procedure ensures that all functions and parameters in the library are utilized, encouraging the \textit{sample\_shape} function to produce outputs that span the observed structures from the validation step.

\begin{table*}[t]
    \centering
    \small
    \setlength{\tabcolsep}{3pt}
    \caption{
    Quantitative evaluation. 
    Comparing different library learning methods along: function usage, semantic consistency, and reconstruction.
    }
    \begin{tabular}{@{}rcccccccccc@{}}
         & \multicolumn{3}{c}{\rule[1.5pt]{9em}{0.5pt} \textbf{\textit{Function Usage}} \rule[1.5pt]{9em}{0.5pt} } 
        & & \multicolumn{3}{c}{\rule[1.5pt]{6em}{0.5pt} \textbf{\textit{Semantics}} \rule[1.5pt]{6em}{0.5pt}} 
        & & \multicolumn{2}{c}{\rule[1.5pt]{3em}{0.5pt} \textbf{\textit{Reconstruction}} \rule[1.5pt]{3em}{0.5pt}}  
        \\
        
        \textbf{Method} & 
        \textbf{\# Fns per Shape}  $\downarrow$ &
        \textbf{\# Fns per Lib}  $\downarrow$ &
        \textbf{Prog Dof}  $\downarrow$ & &
        \textbf{Precision}  $\uparrow$ & 
        \textbf{Recall}  $\uparrow$ & 
        \textbf{F1 score}  $\uparrow$ & &
        \textbf{F-Score (\textit{PC})}  $\uparrow$ &
        \textbf{IoU (\textit{Voxel})}  $\uparrow$ \\
        \midrule
        \llmbaseline &  14.9        & \textbf{5.6} & 63.5          & & 34          & 12          & 18          & & 37.4 & 34.9 \\
        ShapeCoder  &   13.5        & 19.2         & 48.6          & & 25          & \textbf{30} & 27          & & 40.9 &  32.3\\
        \methodname  &  \textbf{9.6} & \textbf{5.6} & \textbf{47.8} & & \textbf{50} & \textbf{30} & \textbf{36} & & \textbf{54.0} & \textbf{50.0} \\
        \bottomrule
    \end{tabular}
    \label{tab:lib_learn_comp}
\end{table*}

\subsection{Training a Recognition Network}
We implement our recognition networks as autoregressive Transformer decoders~\cite{att_is_all}.
These networks use a causal prefix mask to attend to tokens produced by an encoder that consumes an input shape.
They output a program, using the library functions, as a sequence of discrete tokens.
This paradigm supports different visual input modalities by changing the kind of encoder: e.g., a shape represented as a collection of unordered primitives can treat discretized primitive parameters as tokens, 
while a shape represented as a voxel grid can be encoded with a 3D-CNN.
We use the \textit{sample\_shape} function to create paired (visual input, program) data, which we use to train the recognition network with supervised updates (cross-entropy loss on program tokens).
Specifically, we sample a random shape program using the LLM designed procedure, execute the program to produce a collection of primitives, and optionally convert these primitives into unstructured geometry (voxelization or surface sampling).
As the \textit{sample\_shape} functions use randomized calls, they produce an unlimited amount of paired data, so we train in a ‘streaming’ fashion by creating paired data on the fly.

\section{Evaluating Shape Abstraction Libraries}
\label{sec:res_lib_overview}

We evaluate \methodname with experiments over multiple shape modeling domains, split by category: \texttt{chair}, \texttt{table}, \texttt{storage}, \texttt{lamp}, \texttt{faucet}.
For each category, design intent is provided as high-level descriptions of functions that would be useful for this category and
as a set of 20 seed shapes sourced from PartNet~\cite{PartNet}. 
This input is provided to \methodname, which then produces libraries of abstraction functions for each category.
Unless otherwise noted, we use OpenAI's o1-mini as the LLM and gpt-4o as the VLM.

We compare \methodname against alternative abstraction discovery methods that convert this user-provided design intent into a library of programmatic shape abstractions.
\textit{ShapeCoder}~\cite{jones2023shapecoder} is a recent system that achieves state-of-the-art performance for shape abstraction discovery tasks by relying purely on geometric reasoning, without semantic guidance (from any LLM).
We also consider, ~\textit{\llmbaseline}, an alternative version of our approach that ignores the seed set, asking the LLM to implement the library from the function descriptions alone.
We are not aware of any other existing systems that address our problem statement.

\subsection{Library and Function Usage}
\label{sec:res_lib_usage}

To measure how well discovered libraries generalize beyond the seed set, we use recognition networks to infer shape programs that explain validation shapes from PartNet. 
First, we consider the case where each shape is represented as a collection of unordered primitives.
For this reconstruction task (Table~\ref{tab:lib_learn_comp}), we track 
how many function calls were needed to reconstruct each shape (\textit{\#Fns per Shape}) and 
how many degrees of freedom are exposed in the inferred shape programs (\textit{Prog Dof}).
\textit{Prog Dof} is computed as the total number of function calls plus the total number of exposed input parameters in the inferred program, so lower values indicate a simpler interface.
We also report how many functions are included in each library (\textit{\#Fns per Lib}).
From these results, we observe that \methodname offers distinct advantages over both \llmbaseline and ShapeCoder.
It uses fewer function calls for each inferred shape program, exposing a simpler interface.
Compared with \llmbaseline, we find programs that successfully apply the abstractions more often, leading to improved \textit{Prog Dof} values.
Compared with ShapeCoder, \methodname discovers significantly more compact libraries, avoiding function bloat.

\begin{figure}[b!]
\centering
 \includegraphics[width=\linewidth]{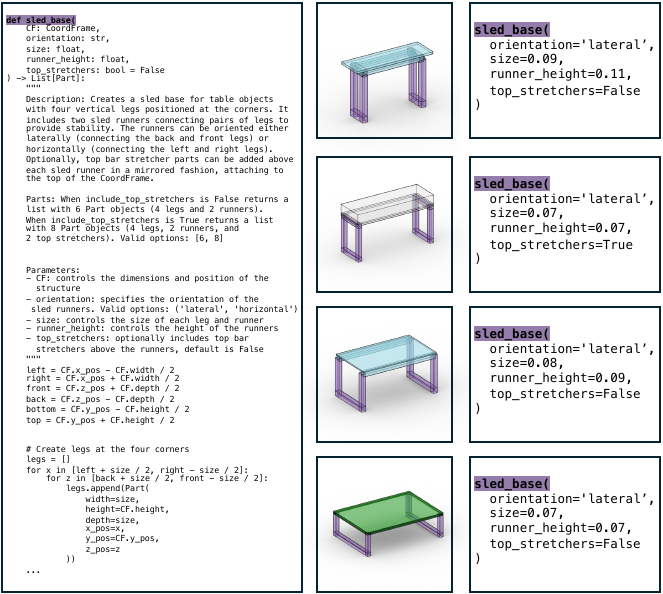}
\caption{
Shape abstractions. 
We show an example of an abstraction function (`sled\_base') produced by \methodname for the \texttt{table} domain. On the left, we provide the validated LLM function implementation. On the right, we demonstrate how different parameterizations of the function are able to cover a family of sub-structures found in validation shapes.} 
\label{fig:qual_libs}
\end{figure}

In Figure~\ref{fig:qual_libs}, we show an example abstraction that \methodname develops, `sled\_base', for the \texttt{table} domain. 
The function exposes a well-defined interface with interpretable parametric handles, such as the height of the runner, or the presence or absence of top stretcher bars.
As indicated, the implementation of `sled\_base' converts these input parameter values into an output part arrangement. 
On the right side of the Figure, we show examples of how different parameterizations of the `sled\_base' function were used to help reconstruct validation shapes.  

\begin{figure*}[t!]
\centering
 \includegraphics[width=\linewidth]{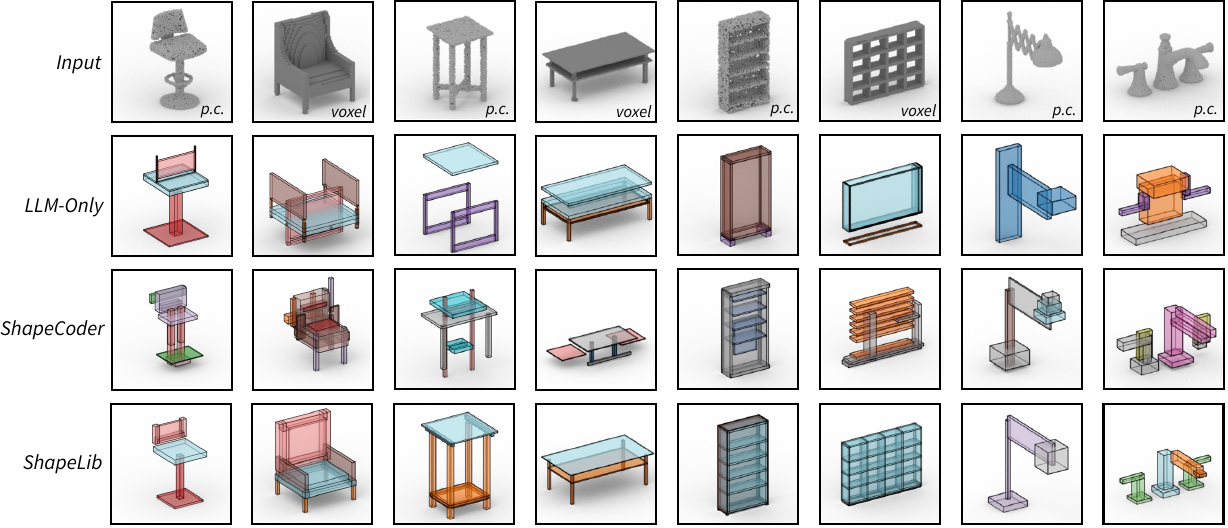}
\caption{
Reconstructing shapes with programs. 
Our recognition networks learn how to reconstruct input shapes (here represented as either point clouds or voxels) with shape programs that use the discovered library functions. Using \methodname abstractions leads to better reconstructions compared with libraries produced by alternative methods. 
We color parts according to the function that produced them.
This highlights the strong semantic alignment of \methodname, both across multiple shapes (light blue seats in columns 1 and 2) and within a single shape (differentiating shelves and frame in column 6).
} 
\label{fig:recon}
\end{figure*}

\subsection{Semantic Alignment}
\label{sec:res_sem_align}

We show examples of inferred programs from validation shapes represented with primitives in
Figure~\ref{fig:prog_comp}.
Compared with ShapeCoder's predictions (left), \methodname's predictions (right) provide a programmatic interface that is easier to understand, as each function and parameter has a semantically relevant name.
Beyond this, ShapeCoder uses multiple abstractions to represent the same type of structure in different reconstructions, which is undesirable. 
In contrast, 
\methodname's functions are applied in a consistent fashion across the entire shape collection.

We quantitatively compare the semantic alignment of these different libraries by using them to perform semantic segmentation.
For each function, we look at the applications made over the seed set, and record the semantic labels of parts that each function explains. 
We then aggregate this information by counting the most commonly covered part labels to produce a simple voting function, which assigns semantic labels when the function is applied.
We evaluate the semantic segmentation performance on fine-grained part labels from PartNet over validation shapes, and report results in the \textit{Semantics} columns of Table~\ref{tab:lib_learn_comp}.
ShapeCoder and~\methodname achieve a similar recall, but~\methodname is twice as precise in its semantic predictions.
\llmbaseline is more precise than ShapeCoder; 
however, without access to seed set exemplars, it cannot find many successful function applications, resulting in poor recall.

\subsection{\methodname method Ablations}
\label{sec:res_method_abl}

We validate our design decisions with a series of ablation experiments over \methodname's components. 
Using the experimental set-up and metrics discussed in the previous section, in Table~\ref{tab:main_app_variants} we report the performance of two ablation conditions of particular note: (i) the LLM model used by \llmbaseline and (ii) how shapes are selected for the seed set.
The following paragraphs analyze the high-level trends from these results.
For a complete description and expanded results for all of our ablation conditions,
including variations of function descriptions and method hyper-parameter settings, we refer readers to Appendix Sections ~\ref{sec:app_res_alternative_abl} and ~\ref{sec:app_res_lib_design}.
Taken together, these results show that while better aligned design intent helps \methodname, the method remains fairly robust to cases where seed sets or function descriptions are only approximately aligned with the target abstractions.

\begin{table}[b!]
    \centering
    \footnotesize
    \caption{Ablations. 
    We compare alternative formulations against \methodname on library learning metrics, demonstrating the importance of seed sets and functional descriptions. For the chair class, we report how functions are used on validation set, and how semantically aligned their applications are.}
    \begin{tabular}{@{}crcccccc@{}}
        \textbf{Condition} &\textbf{Method} & 
        \textbf{\# Fns per Shape}  $\downarrow$ &
        \textbf{Prog Dof}  $\downarrow$ & &
        \textbf{F1 score}  $\uparrow$ \\
        \midrule 
        \textit{Baseline}  & ShapeCoder  &  15.2           &  57.8         & &    31          \\
        \midrule
    \llmbaseline  & o1mini  &  14.6          &  65.6         & &      33          \\
       & claude &  13.5           &  62.5         & &       34          \\
                       & deepseek &  13.4         &  61.4         & &       38          \\
                       & qwen     &  17.2         &  74.2         & &      19          \\
        \midrule
        \textit{Seed Set}  & Random v1     &  12.1           &  57.4         & &      46          \\
                           & Random v2     &  11.2           &  54.9         & &      47          \\
        \midrule        
        \textit{Ours}   & \textbf{\methodname} &  10.9  & \textbf{53.8} & & \textbf{49}  \\
        \bottomrule
    \end{tabular}
    \label{tab:main_app_variants}
\end{table}

\paragraph{\llmbaseline Variations}

By default the \methodname condition uses o1mini as its LLM.
In the \methodname rows in Table~\ref{tab:main_app_variants},
we investigated how library learning performance changes when this LLM is replaced with other frontier LLM models :
qwen-2.5-coder (\textit{qwen}), 
deepseek-r1 (\textit{deepseek}), 
and claude-3.7 (\textit{claude}).
While some of these models do better or worse than others,
we observed none of the LLM variants were able to match the performance of \methodname without guidance from the \textit{seed set}.
Unlike any approach that relies entirely on the prior of an LLM,
\methodname uses the signal from the seed set to help constrain its generated abstractions, so that they model realistic part layouts, and avoid LLM-hallucination failure modes.

\paragraph{Seed set formulation}

\methodname is designed to operate over harmonious system inputs provided by a user: ideally, the shapes in the seed set should match the concepts listed in the function descriptions.
What happens when these design intents are less tightly coupled? 
To explore this question, we consider an alternative seed set selection strategy, where shapes are randomly sub-sampled from a larger collection~\cite{PartNet}.
In Table~\ref{tab:main_app_variants}, we report the results of running \methodname over two seed sets constructed by random sampling under different seeds (\textit{random v1/v2}).

While using randomly constructed seed sets leads to a slight drop in performance, this variant of \methodname still outperforms the best existing alternatives (ShapeCoder or \llmbaseline).
An important consideration for this setting is that while randomly sampled seed sets are likely to contain representative examples for common part structures across a shape collection, they are unlikely to contain examples of less common structures.
A potential failure mode then, is that there would not be seed set shape examples that correspond with certain user provided function descriptions.
Case in point, the \textit{random v1/v2} variants were not able to successfully validate library functions for either the `cantilever base' or `pedestal base' abstractions.
In contrast, when using the user-designed seed set, \methodname is able to find successful implementations for all of the described functions for the chair class.

We expand on this analysis in Appendix Section~\ref{sec:app_res_seed_size}, where we run \methodname over randomly sampled seed sets of varying sizes.
We find that using user-curated seed sets with 20 shapes provides the best trade-off between performance and API cost.
\begin{figure*}[t!]
\centering
 \includegraphics[width=\linewidth]{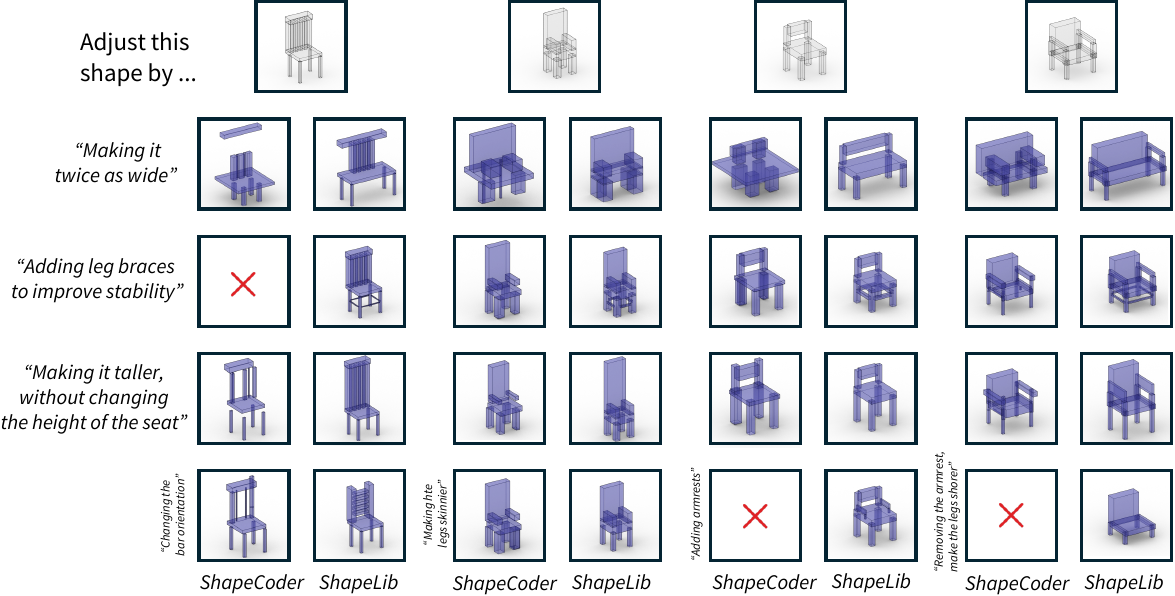}
\caption{
Shape program editability. 
Example comparison conditions from our shape program editing study, where a LLM is tasked with editing a shape (top-row) represented as either a \methodname or ShapeCoder program with respect to an edit request (left column). 
The programmatic interface exposed when using \methodname abstractions is better aligned with semantics and easier for the LLM to understand, so we consistently observe that LLMs produce more semantically consistent and accurate shape edits when operating over \methodname programs.
LLM predictions that caused a run-time error are marked with `X'.
} 
\label{fig:edit_prog}
\end{figure*}

\section{\methodname Applications and Extensions}
\label{sec:res_app_overview}

We have shown that \methodname generates libraries of shape abstractions that effectively generalize and well-align with semantics, but what can we do with these abstraction functions?
To this end, we explore a range of downstream applications supported by \methodname.
First, in Section~\ref{sec:res_recon} we show how our recognition networks can be used to find shape programs with abstractions that represent unstructured input geometry.
Then, in Section~\ref{sec:res_edit} we demonstrate that our programmatic interface unlocks editing capabilities over both shape structures and textured meshes.
Finally, in Section~\ref{sec:res_gen} we investigate how \methodname's functions can integrate with generative workflows, again handling structural and detailed geometric representations. 

\subsection{Reconstructing Unstructured Geometry}
\label{sec:res_recon}

In the previous section, we investigated the behavior of our structured recognition networks that consume shapes represented as primitive soups, but what about unstructured geometry?
As discussed in Section~\ref{sec:lib_usage}, we can generalize our recognition networks for this setting, by simply replacing the primitive encoder with either a point cloud or a voxel encoder.
With this network we can then solve visual program induction tasks, finding shape programs that reconstruct shapes from outside of the seed set.

\subsubsection{Comparing Library Variations}

We have found that \methodname's library of programmatic abstraction functions finds more consistent and semantic function applications compared with libraries produced by alternative methods.
Are \methodname's abstractions similarly beneficial for this reconstruction task? 
We design an experiment to test this question, by training versions of our recognition networks on function libraries produced by \methodname, ShapeCoder, and \llmbaseline.
We report results for this experiment in the \textit{Reconstruction} rows of Table~\ref{tab:lib_learn_comp}.
For the point cloud to program task, we sample a point cloud from the abstracted cuboid outputs, and report the F-score~\cite{TanksAndTemples}.
For the voxel to program task, we convert the program's execution into a voxel field, and report IoU.
We find that using \methodname's abstractions infers more accurate reconstructions from these unstructured inputs, compared with the function libraries produced by ShapeCoder or \llmbaseline.
We visualize some qualitative comparisons of this experiment in Figure~\ref{fig:recon}. 
Beyond more faithful reconstructions, again we observe that \methodname's function applications are more semantically correlated, even in this unstructured input setting.

\begin{figure*}[t!]
\centering
 \includegraphics[width=\linewidth]{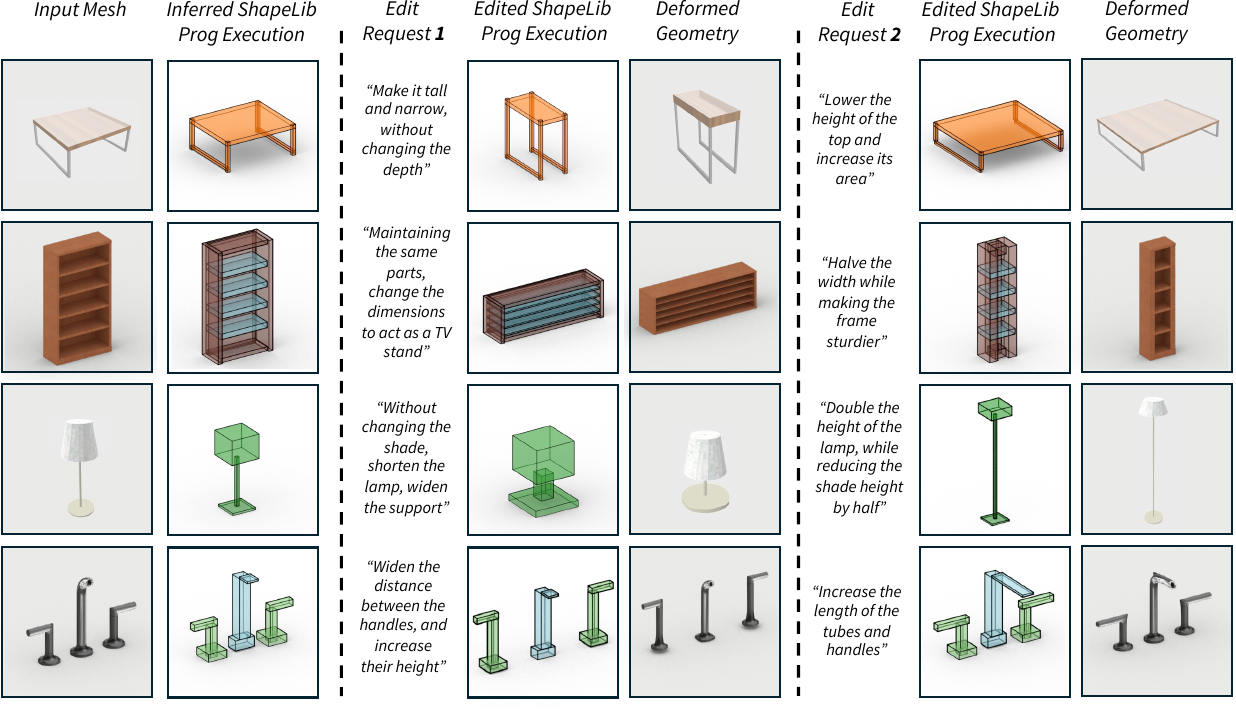}
\caption{
Editing textured meshes. 
Demonstrations of how \methodname can be used to edit textured meshes. Given an input mesh (left column), we use our recognition networks to infer a \methodname program. This program can then be edited with an LLM, here we show two potential edit requests and the execution result of the edited program. 
We use the program edit to guide a cage-based deformation of the input geometry, using the cuboids as the cage.
} 
\label{fig:edit_geom}
\end{figure*}

\subsubsection{Comparing Recognition Network Variations}
\label{sec:res_recon_variations}

As described in Section~\ref{sec:lib_usage}, we train our recognition networks from scratch, sampling paired (\textit{shape}, \textit{program}) data from LLM authored \textit{sample\_shape} functions. 
However, due to the success of frontier VLMs across a diverse range of visual reasoning tasks, one may wonder: could these models be used to solve this task directly?
We design an experiment to evaluate this question. 
We provide a LLM with vision capabilities (gpt-4o) with \methodname's discovered library of abstraction functions, along with in-context examples of how to represent seed set shapes with these abstraction functions. 
From this conditioning, we then ask it to author a program that would reconstruct a new validation shape (we visualize one example output in Figure~\ref{fig:llm_motiv}). 
As VLMs cannot naturally take in voxels or point clouds inputs, we instead represent each shape as either a list of primitives or as a rendered image. 
We report results for this experiment, over a set of validation chair shapes in Table~\ref{tab:vlm_Recon}.
This complex task of visual program induction, finding a program that reconstructs an input shape, is too hard of a problem for frontier LLMs to solve directly.
This justifies the framing of our library-and-task-specific recognition networks, that learn to specialize much smaller networks to offer greatly improved reconstruction performance.

By default, the \textit{sample\_shape} functions that we use go through an iterative refinement procedure as described in Section~\ref{sec:lib_usage}.
We demonstrate the benefit of this procedure by comparing it against two alternatives: (i) using the first version of the \textit{sample\_shape} function produced by the LLM or (ii) independently sampling, then aggregating, k versions of the \textit{sample\_shape} function.
We find that our iterative procedure produces training data that leads to better recognition networks, as measured by reconstruction performance (see Appendix Section~\ref{sec:app_res_rec_net} for details).

\begin{table}[t!]
    \centering
    \caption{Recognition network alternatives. 
    Comparing VLMs against library specific recognition networks for reconstructing shapes with programs.}
\begin{tabular}{@{}rl@{\hspace{2.5em}}c@{}}
        \toprule
        \textbf{Method} & \textbf{Input Modality} & \textbf{IoU $\uparrow$} \\
        \midrule
        gpt-4o & image & 18.2 \\
        gpt-4o & primitives & 38.4 \\
        \methodname & voxels & \textbf{62.2} \\
        \bottomrule
    \end{tabular}
    \label{tab:vlm_Recon}
\end{table}

\subsection{Editing Shape Programs}
\label{sec:res_edit}

\subsubsection{Editing Shape Programs with LLMs.}
A good library of programmatic abstractions should expose an interface that is easy-to-use and maintains shape plausibility under manipulation.
We evaluate these properties by comparing how well an agent (i.e., an LLM) can edit shape programs in a goal-directed fashion.
With libraries produced by either \methodname or ShapeCoder, we first use recognition networks to find shape programs that reconstruct validation shapes.
Then, after designing a series of shape edit requests in natural language, we can ask an LLM to edit the text of the shape program to meet the request (i.e., change function parameters and how functions are used, as depicted in Figure~\ref{fig:teaser}).

We provided \textit{o1mini} with the fully implemented function library for both~\methodname and ShapeCoder conditions. 
To evaluate performance, we designed a two alternative forced choice perceptual study.
We recruited 13 participants, who made 25 judgments each, over 100 possible shape edit comparisons.
We asked each participant to make two judgments: (i) which manipulated shape was more plausible;
and (ii) which edit better matched the input edit request.
The results of this perceptual study (Table~\ref{tab:llm_edit}) provide further support of \methodname advantages.
Our library of shape abstraction functions provides an easy-to-use interface, leading to edits that more often match the edit intent, while at the same time maintaining better shape plausibility under parameter variations.
We show qualitative demonstrations of these edits in Figure~\ref{fig:edit_prog}, and observe higher semantic alignment of LLM edits using \methodname programs.
\begin{table}[h!]
    \centering
    \small
    \caption{
    Perceptual study. 
    Results of our perceptual study evaluating edits made by an LLM to programs that use shape abstraction libraries.  
    We report judgments along two axes: shape plausibility and match to edit intent. 
    }
    \begin{tabular}{@{}rcc@{}}
        \toprule
                & \textbf{More Plausible(\%)} & \textbf{Better Matches Intent (\%)} \\
        \midrule
        vs. ShapeCoder       &       75\%       &    73\%                             \\ 
        \bottomrule
    \end{tabular}
    \label{tab:llm_edit}

\end{table}

\subsubsection{Deforming Geometry with Program Edits.}
LLMs can successfully edit abstracted shape representations through program modifications, but what about edits to detailed geometry?
We show examples in Figure~\ref{fig:edit_geom} of how \methodname can support text-driven mesh edits.
In this proposed workflow, a mesh is first converted into a point cloud, which we pass into our recognition networks to find a reconstructing program.

Providing this shape program as input, we ask an LLM to modify the program with respect to an edit request.
Executing the original and edited program gives us a starting cuboid layout and a target cuboid layout.
From this start and end state, we can use a deformation scheme to modify the original positions of the mesh vertices with respect to the edited program.

We find that a relatively lightweight cage-based deformation scheme supports a range of interesting edits.
For each mesh vertex, we find the local position of the vertex within each cuboid in the starting layout.
We assign a weight for this vertex to each cuboid in the layout.
If the vertex is inside a single cuboid, all of its weight goes to that cuboid.
Otherwise, if the vertex is inside multiple cuboids, or no cuboids, we assign the weight as a distribution over all of the cuboids, determined by soft blend using inverse point-to-cuboid-surface distance.
With this weighting scheme, given a new primitive layout (e.g. from a program edit and execution), we can produce updated vertex positions.
We calculate the local position of each vertex with respect to each new cuboid placement, and then average these positions with respect to the cuboid weighting distribution.
As shown in Figure~\ref{fig:edit_geom}, this workflow is able to produce compelling shape edits by leveraging the benefits of \methodname.

\begin{figure*}[t!]
\centering
 \includegraphics[width=\linewidth]{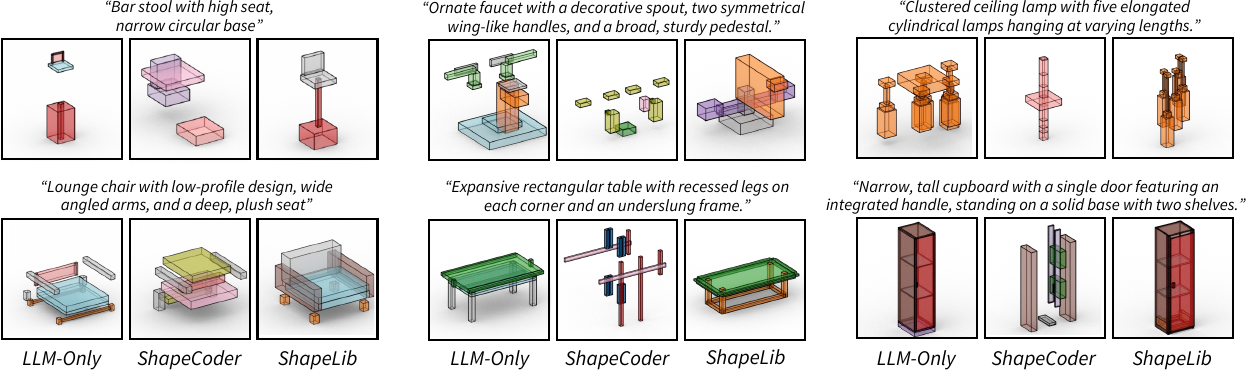}
\caption{
LLMs using shape abstractions. 
We evaluate the ability for LLMs to use different abstraction libraries to generate shape programs that correspond with input text prompts. We find that LLMs are more successful in this task when presented with \methodname (ours) functions.
} 
\label{fig:prog_gen}
\end{figure*}

\begin{figure*}[t!]
\centering
 \includegraphics[width=\linewidth]{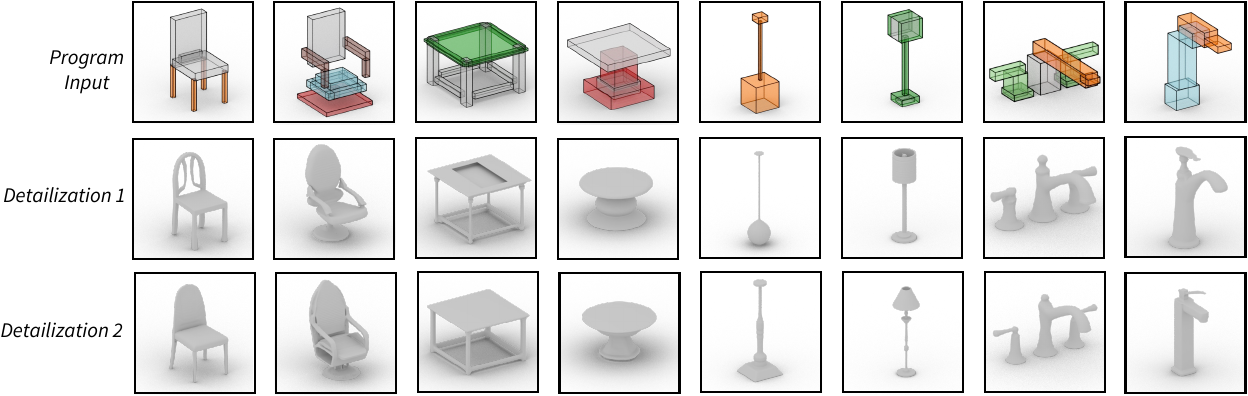}
\caption{
Geometric detailization of shape abstractions. 
 We demonstrate how \methodname program executions (top row) can be converted into detailed geometry with a structure-conditioned 3D generative model. 
 In the middle and bottom rows, we show variations for how a single input structure can be stylized.
} 
\label{fig:geom_gen}
\end{figure*}

\subsection{Shape Generation}
\label{sec:res_gen}

\subsubsection{Structure Generation with Shape Programs}
Beyond shape editing, we explore how \methodname can benefit generation tasks.
To generate new structured shape representations, we can synthesize new shape programs, but who will author them?
We design an experiment that uses LLMs to convert text descriptions into shape programs.
We provide an LLM with a library of shape abstraction functions, and in-context examples of how text descriptions map to shape programs, sourced from the seed set.
Then, given a new text description, we can ask the LLM to author a new shape program.
Given a starting set of 10 shape descriptions, we prompt an LLM through ideating 50 new descriptions, designed to span common structural variations across a category. 
With this set, we compare \methodname against ShapeCoder and \llmbaseline,  generating 2 shape programs per description (producing 500 shape programs per method in total).

We report results of this experiment in Table~\ref{tab:gen_struct}.
We can evaluate the distributional similarity of the LLM generated structures against structures from PartNet validation shapes, which act as a reference set.
For Frechet PointNet++ Distance (\textit{FPD}) and Kernel PointNet++ Distance (\textit{KPD}) we use a feature space of a pretrained ShapeNet classification network.
Minimum matching distance (\textit{MMD)} reports the average minimum Chamfer distance of each member of the reference set to any member of the generated set.
We also show qualitative examples of this text-to-program generation task in Figure~\ref{fig:prog_gen}.
Overall, we find that LLMs generate better shape structures when using \methodname libraries, as our semantically aligned abstractions expose an easy-to-use programmatic interface.

\begin{table}[b!]
    \centering
    \small
    \caption{
    LLM shape program generation. We measure the distributional similarity of LLM produced shape structures against validation set shape structures (averaged over each category).}
    \begin{tabular}{@{}rcccc@{}}
        \toprule
        \textbf{Method} 
        & \textbf{FPD} $\downarrow$ 
        & \textbf{KPD} $\downarrow$ 
        & \textbf{MMD} $\downarrow$ 
       \\
        \midrule
         \llmbaseline  & 2219 & 717 & 0.100   \\
         ShapeCoder    & 449 & 31 & 0.091  \\
         ShapeLib      & \textbf{278} & \textbf{26} & \textbf{0.079} \\
        \bottomrule
    \end{tabular}
    \label{tab:gen_struct}
\end{table}

\subsubsection{Structure-conditioned Shape Detailization}

What if we want to convert these abstracted shape structures into more detailed geometry?
For this task, we use a structure-conditioned detailization module that consumes a set of primitives and outputs a mesh with geometric details, as shown in Figure~\ref{fig:teaser}.
While a few existing systems have been designed for this purpose~\cite{sella2024spice, chen2021decor}, they do not scale as well to the structurally complex primitive layouts produced by \methodname programs.
Therefore, we propose a lightweight alternative where we adapt a recent state-of-the-art 3D generative model to condition its predictions on an input set of primitives.

\paragraph{Structure-aware Shape Generation}

In our adaptation, we start with the pretrained \textit{Cube} model~\cite{roblox2025cube}.
The base \textit{Cube} model uses a decoder only Transformer to learn how to map CLIP embeddings of shape descriptions to sequences of discrete shape tokens, that can be decoded into a 3D mesh.
To make the model structure-aware, we remove this CLIP conditioning, and instead add in a box encoder.
This box encoder consumes a layout of cuboid primitives, lifts each primitive to a higher dimension with a linear layer, and then combines these features through self-attention layers.
We finetune the network with paired cuboid layouts and meshes with geometric details, sourcing this data from PartNet~\cite{PartNet}, and training in a per-category fashion.

\paragraph{Integration with \methodname}

To integrate this network with \methodname programs, we take \methodname programs and execute them to get a layout of cuboids.
We pass this as input into the network, which then autoregressively generates a sequence of discrete shape tokens, which we can pass into the frozen decoder to produce a 3D mesh.
We sample tokens from the transformer with Top-P sampling of 90\%, so we get variations in the style, and quality, of the detailizations.
As shown in Figure ~\ref{fig:geom_gen}, this simple approach performs quite well.
We can convert structured \methodname programs into meshes with interesting geometric details.
This detailization process largely respects the input layout, and demonstrates multiple plausible variations starting from the same part conditioning input.

\section{Discussion}
\label{sec:discussion}
We have presented \methodname as the first method that uses LLM priors to produce a library of programmatic 3D shape abstractions that \emph{generalize} over shape collections, expose \emph{interpretable} parameters, and maintain \emph{plausible} outputs under manipulation.
We make use of two forms of complementary design intent, a seed set of example shapes and descriptions of functions to include in the library, to create shape programs that are compact and semantically well-aligned, supporting downstream applications like reconstruction, editing, and generation.

\subsection{Failure Modes and Limitations}

\subsubsection{Library Development with LLMs}

As with other shape analysis methods that rely on LLMs, \methodname needs to deal with the fact that LLM outputs can introduce errors. 
These errors can be (1) \textit{bugs} (formatting or run-time errors) or (2) \textit{semantic errors} (valid code that produces geometry that does not represent actual shapes).
ShapeLib uses the geometric validation step (d) to detect and handle both of these error types.
In steps (b) and (c) of our pipeline, the LLM produces many application proposals ($K_A$) and many implementation proposals ($K_I$), resulting in $K_I * K_A$ total proposals. 
Proposals with poor reconstruction of patterns derived from the seed shapes, due to either \textit{bugs} or \textit{semantic errors}, get filtered out.

\begin{figure*}[t!]
\centering
 \includegraphics[width=\linewidth]{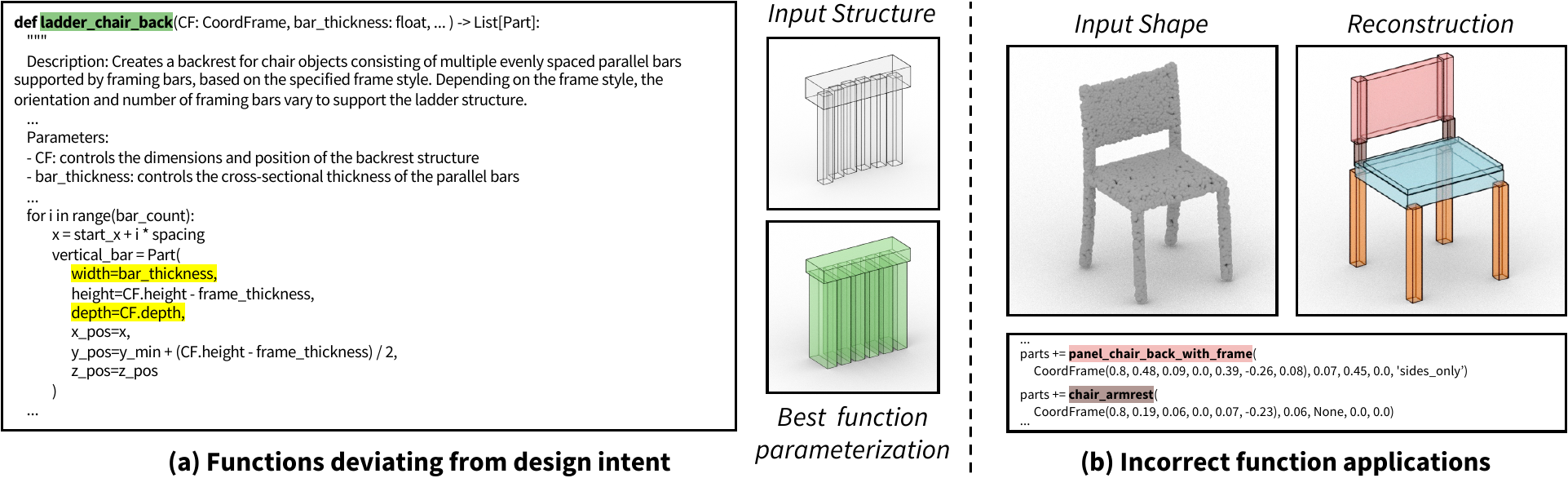}
\caption{
Failure modes of \methodname.
\textit{(left):} LLM proposals with semantic errors can still be added into the library, as long as these errors still `well-reconstruct' seed set examples. In this case, the depth parameter should also be set to `bar\_thickness'. 
As a result, this function cannot find good parameterizations for some input shapes that have a ladder back structure.
\textit{(right):} Recognition networks can apply the wrong abstractions during reconstruction. 
In this case, an extraneous `chair\_armrest' function is used in-place of two back supports, to represent geometry that could have instead been explained entirely with the `panel\_chair\_back\_with\_frame' function.
As a result, the reconstructing program is longer, and the exposed interface contains semantic inconsistencies.
} 
\label{fig:fail_mode}
\end{figure*}

\paragraph{Buggy LLM Productions}

Before validation, we find that while \textit{bugs} are not prevalent, they are also not uncommon, occurring in 5-10\% of LLM proposals. 
After the validation stage, we find that the library functions rarely contain \textit{bugs}, as evidenced by the fact that <1\% of all synthetic data sampling calls produce a run-time error.
In Appendix Table~\ref{tab:method_opts} we show that when setting $K_A$=1 and $K_I$=1, we get \textit{semantic errors} rather than \textit{bugs}; if the LLM fails to produce any proposals for a function that validate on the seed set, we remove the function from the library.

\paragraph{Design Intent Mismatch}

\methodname assumes that the guidance signals provided from the user are complementary.
In Section~\ref{sec:res_method_abl} we explored what happens when user-specified design intents are replaced with automatic variations that break this assumption slightly.
Taking this a step further, in extreme cases, there might be no overlap between the two input modalities, if none of the function descriptions have corresponding patterns in the seed set shapes.
For this setting, due to \methodname's reliance on seed-set based validation, the failure mode is that no abstraction functions would get successfully added into the library. 
While we did not observe this issue in our experiments, this problem could likely be mitigated by introducing an iterative loop to \methodname where the design intent is modified (by a user or automatic process) whenever a function fails to validate on the first pass.

\paragraph{Function Deviation from Design Intent}

A limitation of our validation stage is that it relies on an error metric to check proposal correctness (see Section~\ref{sec:lib_validation}), where this error metric uses a soft-threshold to determine if a predicted layout `well-reconstructs' a layout from the seed set.
With too low of a matching tolerance, the LLM would need to predict exactly correct function parameterizations.
However, by allowing a small amount of tolerance for geometric error, there can be cases where function implementations with small semantic mistakes are included in our libraries.
We show such an example in Figure~\ref{fig:fail_mode} (left), for the `ladder\_back' function. 
The input design intent indicates that each bar should have a square-cross section (equal width and depth).
The LLM produced solution instead sets the depth of each bar equal to the depth of the entire back part -- this allows the depth and width of each bar to deviate from one another, invalidating the desired constraint.
Despite the error, this implementation of the abstraction function still produces outputs that `well-reconstruct' the seed set shapes under our error metric, and prove useful for downstream reconstruction, generation and editing applications.
To better understand the prevalence of this issue, we provide an audit of \methodname validated libraries in Section~\ref{sec:app_res_fn_audit}.
Overall, while \methodname does a much better job of implementing a library of abstraction functions that respect the input design intent compared with \llmbaseline or ShapeCoder, this failure mode shows that there are further opportunities to improve our discovered functions.

\subsubsection{Recognition Networks}

Beyond library development, our recognition networks can introduce another source of error in terms of how the abstraction functions are applied during reconstruction. 
We show one such failure mode in Figure~\ref{fig:fail_mode} (right). 
In this case, a `chair\_armrest' function call has been inappropriately used to represent the supporting parts of the chair back. 
Though the prediction made by our recognition network has a low reconstruction error, this program exposes an interface that might cause confusion in downstream applications.
As our recognition networks train without any human labeled data, only using (program, layout) pairs produced by LLM authored synthetic data generators, some of these semantic misapplications are to be expected.
So, while we do observe that our reconstruction process is both more accurate and has better semantic alignment compared with existing alternatives, there remain opportunities to develop better architectures and training schemes to minimize this type of failure mode.

Our current recognition network setup also assumes clean point clouds sampled from full geometry, rather than raw scanned point clouds with occlusion, sensor noise, or irregular density. Extending \methodname to handle such inputs sourced from 3D scans would likely require combining our pipeline with methods that first map partial or degraded scans into a cleaned-up version that better matches our synthetic data generation setup.

\subsection{Future Work}

\methodname assumes that users have enough domain-knowledge to provide useful design intent to our system in two forms: text descriptions and a seed set of shapes. 
This framing provides the user with greater control over how the library of abstractions is implemented.
Moreover, when compared with the current best alternative for producing high-quality shape abstraction functions, which requires manually writing thousands of lines of code, using \methodname takes significantly less time and effort.
Choosing a small set of exemplar shapes from a larger repository of part-decomposed shapes is fast and does not require much expertise.
Specifying function descriptions can require non-negligible user effort, but the total time consumed by this process is a fraction of that required when implementing each function by hand.
In fact, this input modality is what allows \methodname to discover libraries of semantically-aligned functions, as demonstrated by the fact that the best library learning methods that don’t use these descriptions (ShapeCoder), produce hard-to-use program interfaces (see Figure~\ref{fig:prog_comp}). 
The cost of our system design is that for novice users, who may lack clear design goals, it may be hard to develop good function descriptions.
One possible way to overcome this issue might be to turn to the LLM to help with the automation of eliciting design intent.
For instance, one could treat the LLM as a hypothesis generator for function descriptions, which would then likely need to be validated, perhaps by including the user `in-the-loop'.
Relatedly, another modality that may prove useful in conveying design intent, or guiding LLMs towards producing good libraries of programmatic abstractions, might come through the presentation of examples of expert-crafted procedural models~\cite{infinigen2024indoors}.

At present, \methodname's stages are run once in a single pass to produce an output library from an input design.
A natural next step would be to convert this static process into a dynamic iterative loop, where the design intent is modified on the fly to either expand library coverage or provide refined guidance for hard-to-implement abstractions.
An interesting extension in this space would be to iterate on the design intent until the seed set meets some `well-covered' criteria under the discovered library, where uncovered structures could serve as signals for which abstractions or descriptions should be added or revised.
In such an iterative formulation, one could even imagine a user participating more explicitly in the loop to update their design intent: reviewing failed validations, uncovered structures, or function implementations that don't match their intended semantics.
A related extension would be to augment our current iterative refinement of the \textit{sample\_shape} function with feedback that targets semantic correctness in addition to geometric coverage.

Beyond such workflow changes, \methodname's framing also suggests a number of broader modeling extensions.
For instance, one could extend beyond single-category libraries toward a more general `meta-category' framing, where a shared function library supports a diverse, but related set of categories (e.g. a single library supporting many types of furniture objects, such as chairs, tables, cabinets, etc). 
Another natural extension would be to move beyond independent mid-level abstractions toward explicitly hierarchical abstractions whose concepts can build on, and reference, one another.
While \methodname's current abstractions effectively capture static geometries, how can we make them `physics-aware’ to support dynamic interactions? 
One possible approach is to integrate affordance priors (e.g., supporting relations, joint types, etc.) derived from an LLM or specialized module, together with functional tags and hints provided by the user as an additional form of design intent.
Finally, continued iteration on structured and style-guided detailization modules would be important to improve the coupling between high-level structure and geometric output. 
If this can be achieved while explicitly disentangling `style', it would help bring us closer towards fully automatic, high-quality procedural model creation.

\bibliographystyle{ACM-Reference-Format}
\bibliography{main}

\clearpage
\appendix
\appendix

\section*{Appendix Table of Contents}

\noindent\textbf{Appendix~\ref{sec:app_res}. Additional Results}

\begin{itemize}
\item Section~\ref{sec:app_res_alternative_abl}. ShapeLib Alternative Ablations
\item Section~\ref{sec:app_res_lib_design}. ShapeLib Library Design Ablations
\item Section~\ref{sec:app_res_rec_net}. ShapeLib Recognition Network Ablations
\item Section~\ref{sec:app_res_param_fit}. Accuracy of LLM-Predicted Fn Applications
\item Section~\ref{sec:app_res_seed_size}. Varying Seed Set Size
\item Section~\ref{sec:app_res_fn_audit}. Audit of Design Intent Deviation in Validated Libraries
\end{itemize}

\noindent\textbf{Appendix~\ref{sec:app_method}. Additional Method Details}

\begin{itemize}
\item Section~\ref{sec:app_lib_design}. Library Design
\item Section~\ref{sec:app_rec_net}. Recognition Networks
\item Section~\ref{sec:app_synth_data_sampler}. Synthetic Data Sampler
\item Section~\ref{sec:app_gen_edit_apps}. Generation and Editing Applications
\end{itemize}

\noindent\textbf{Appendix~\ref{sec:app_exp}. Additional Experiment Details}

\begin{itemize}
\item Section~\ref{sec:exp_cost_time}. Cost and Timing
\item Section~\ref{sec:exp_data}. Data
\item Section~\ref{sec:exp_shapecoder}. ShapeCoder
\item Section~\ref{sec:exp_llm_only}. \llmbaseline
\item Section~\ref{sec:exp_llm_shape_edit}. LLM Shape Editing
\item Section~\ref{sec:exp_llm_shape_prog_gen}. LLM Shape Program Generation
\item Section~\ref{sec:exp_vlm_details}. VLMs for visual program induction
\end{itemize}

\noindent\textbf{Appendix~\ref{sec:app_prompts}. Prompt Templates}

\noindent\textbf{Appendix~\ref{sec:app_design_intent}. Example Design Intent}

We provide additional results in Section~\ref{sec:app_res},
additional method details in Section~\ref{sec:app_method}, 
additional experiment details in Section~\ref{sec:app_exp},
representative prompt templates in Section~\ref{sec:app_prompts}, and examples of user-provided design intent in Section~\ref{sec:app_design_intent}.

\section{Additional Results}
\label{sec:app_res}

\subsection{ShapeLib Alternative Ablations}
\label{sec:app_res_alternative_abl}

\begin{table*}[t]
    \centering
    \caption{Ablation experiment comparing how variations of function descriptions used by \methodname affect library learning metrics. For the chair class, we report how functions are used on validation set shape, and how semantically aligned their applications are.}
    \begin{tabular}{@{}crcccccccc@{}}
         & & \multicolumn{2}{c}{\rule[1.5pt]{4em}{0.5pt} \textbf{\textit{Fn Usage}} \rule[1.5pt]{4em}{0.5pt} } 
        & & \multicolumn{3}{c}{\rule[1.5pt]{6em}{0.5pt} \textbf{\textit{Semantics}} \rule[1.5pt]{6em}{0.5pt}}  \\
        \textbf{Condition} &\textbf{Method} & 
        \textbf{\# Fns per Shape}  $\downarrow$ &
        \textbf{Prog Dof}  $\downarrow$ & &
        \textbf{Precision}  $\uparrow$ & 
        \textbf{Recall}  $\uparrow$ & 
        \textbf{F1 score}  $\uparrow$ \\
        \midrule 
        \textit{Baseline}  & ShapeCoder  &  15.2           &  57.8         & &   28        &     34     &  31          \\

        \midrule        
        \textit{Function}  & LLM Reword v1     &  11.1         &  59.0            & &   64      &     36     &  46          \\
        \textit{descriptions}                   & LLM Reword v2     &  11.5         &  58.6            & &   57       &    27     &  37          \\
                           & LLM Generate    & \textbf{10.8}         &  67.8            & &   55       &    32     &  41          \\
        \midrule        
        \textit{Ours}   & \textbf{\methodname} &  10.9  & \textbf{53.8} & & \textbf{65} & \textbf{39} & \textbf{49}  \\
        \bottomrule
    \end{tabular}
    \label{tab:supp_app_variants}
\end{table*}

We expand on the ablation experiments discussed in Section~\ref{sec:res_method_abl}.
\methodname takes two forms of design intent (seed set and function descriptions) that are thought of, and provided by, a user, who has a particular modeling goal in mind.
These descriptions guide the LLM in how to design the interface of the library, which is how the user will interact with the functions when they have been implemented.
We explored how sensitive \methodname is to the seed set shapes in Table~\ref{tab:main_app_variants}, but what about the function descriptions?
We run experiments on the chair category, asking each method to discover libraries of abstraction functions.
We report results of this experiment in Table~\ref{tab:supp_app_variants}, using the Function usage and the semantic consistency metrics from section~\ref{sec:res_lib_usage}.
This involves first generating a library, then generating a program sampler from the LLM, and finally training a recognition network from this sampler that conditions on a collection of cuboids and predicts a shape program that uses abstraction functions.

While the semantic meanings of the function descriptions are clearly critical (e.g. what functions should be added, what types of parameters should be exposed), one might also wonder how sensitive \methodname is to the exact format of these descriptions.
To evaluate this sensitivity we ask an LLM to reword the chair set function descriptions, maintaining the semantic meaning while rephrasing each description independently. 
With this new set of function descriptions, we run \methodname, keeping the seed set fixed, these become the \textit{LLM Reword v1} and \textit{LLM Reword v2} rows in Table~\ref{tab:supp_app_variants}, under the \textit{function descriptions} conditioning heading.
Note that like in the seed set variation experiments, we report multiple runs to test sensitivity. 
This version shows slightly higher performance degradation compared with the seed set variation experiments, but offers performance that is still well above the baselines.
We also consider a more ambitious variation, where we provide the LLM with just a function name, e.g. `ladder chair back', and then ask it to produce the function description (given an in-context example from another category).
We call this version \textit{LLM Generate}.
While the \textit{Prog Dof} metric is dramatically hurt (because the LLM does not design the interfaces to try to minimize exposed degrees of freedom), this variant actually achieves the best performance on the \textit{Fns per Shape} metric, and has competitive semantic consistency performance.

\subsection{ShapeLib Library Design Ablations}
\label{sec:app_res_lib_design}

\begin{table*}[t]
    \centering
    \caption{
    Ablation experiment on parameters used in \methodname's library learning procedure. 
    For chair category, we report how well the libraries represent shapes in the seed set.
    \methodname default settings, in the bottom row, outperform alternative parameter settings, justifying our design decisions.}
    \begin{tabular}{@{}ccccc|cccc@{}}
        \toprule
        \textbf{$K_A$} & \textbf{$K_I$} & \textbf{EPS} & \textbf{ICE} & \textbf{Impl LLM} & \textbf{Obj} $\downarrow$ & \textbf{Valid Fn Apps per Shape} $\uparrow$ & \textbf{Uncovered Parts per Shape} $\downarrow$ \\
        \midrule
        1 & 1 & \checkmark & \textit{param mask} & \textit{o1mini} & 48.5 & 1.7 & 4.1 \\
        1 & 4 & \checkmark & \textit{param mask} & \textit{o1mini} & 47.1 & 1.9 & 3.6 \\
        5 & 1 & \checkmark & \textit{param mask} & \textit{o1mini} & 41.9 & 2.0 & 2.6 \\
        5 & 4 & $x$ & \textit{param mask} & \textit{o1mini} & 44.0 & 2.0 & 3.0 \\
        5 & 4 & \checkmark & \textit{none} & \textit{o1mini} & 45.1 & 2.0 & 3.2 \\
        5 & 4 & \checkmark & \textit{full params} & \textit{o1mini} & 43.6 & 2.0 & 3.0 \\
        5 & 4 & \checkmark & \textit{param mask} & \textit{gpt4o} & 43.6 & 2.1 & 2.9 \\
        \midrule
        5 & 4 & \checkmark & param mask & o1mini & \textbf{40.7} & \textbf{2.2} & \textbf{2.2} \\
        \bottomrule
    \end{tabular}
    \label{tab:method_opts}
\end{table*}

We ablate different parts of the pipeline involved in designing the library of abstraction functions in 
Table~\ref{tab:method_opts}.
In this experiment, we use the chair category, and modify different hyper-parameters of the method.
$K_A$ is number of input-output applications proposed per shape.
$K_I$ is number of proposed implementations per function.
\textit{EPS} modifies whether we consider the expanded parameter set (see Section~\ref{sec:app_lib_val}) or just the LLM produced input-output pairs.
\textit{ICE} modifies how the in context input-output
examples are formulated for the LLM implementation stage.
By default, we  mask parameters with `?' values (\textit{param mask}). 
We consider alternatives when no input-output examples are provided (\textit{none)} or where all input-output examples have the LLM predicted parameters not masked out (\textit{full params}).
We also consider a variant where we change the LLM model used for the function implementation proposal step, from \textit{o1mini} to \textit{gpt4o}.

Under variations of these parameters, we discover shape program abstraction libraries.
We then record how well these libraries can represent the seed set shapes.
These metrics are:
the objective value (\textit{Obj}) achieved by the resulting programs, where the goal is to have a low compression-based objective (see Section~\ref{sec:app_lib_learn_obj}); 
the number of successfully validated function applications made per shape (\textit{Valid Fn Apps per Shape}) -- it is good when this metric is high, as it means that the LLM has produced functions that can be used frequently over the seed set;
the average number of parts in each seed set that are not explained by function calls (\textit{Uncovered Parts per Shape}) -- we want this number to be low, so that most of the parts in each seed set shape are covered by the library logic .
Our default settings, in the last row, achieves the best performance for these metrics, justifying our design decisions.

\subsection{ShapeLib Recognition Network Ablations}
\label{sec:app_res_rec_net}

\begin{table}[ht]
    \centering
    \small
    \caption{
    We compare recognition network performance on point cloud to shape program reconstruction task for the chair category, under different ways of creating the \textit{sample\_shape} function.
    }
    \begin{tabular}{@{}rcc@{}}
        \toprule
        \textbf{Method}    & \textbf{CD} $\downarrow$ & \textbf{F-Score} $\uparrow$ \\
        \midrule
        \textit{single sampler} &   0.046   &  46.5    \\
        \textit{indep sampler}  &   0.043   &  49.8   \\
        \textit{iter sampler}   &   \textbf{0.039}   &  \textbf{51.8}   \\
        \bottomrule
    \end{tabular}
    \label{tab:app_vpi_net}
\end{table}

We use an LLM generated  \textit{sample\_shape} function to produce training data for our recognition network.
We use an iterative error correction procedure to iterate on the design of this sampler (see Section~\ref{sec:app_synth_data_sampler})

We justify this design choice with a small ablation.
For the chairs category, we take our point cloud to shape program recognition network, and train multiple versions.
In the default setting, we take the first authored \textit{sample\_shape} function, and produce two improvements with our LLM prompting workflow.
We then train on data produced by all three functions (\textit{iter sampler}).
We consider two alternatives.
As one option, we just take the first \textit{sample\_shape} function produced by the LLM, without iterative correction (\textit{single sampler}).
As another option, we create three independent \textit{sample\_shape} functions, by using the same prompt three times, and train on data produced by all three functions (\textit{indep sampler}).

\begin{figure}[h]
\centering
 \includegraphics[width=\linewidth]{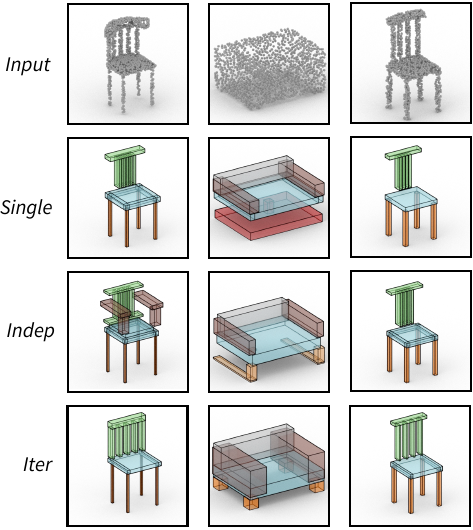}
\caption{
We show representative program reconstructions for the recognition network settings compared in Table~\ref{tab:app_vpi_net}: training on a single initial sampler (\textit{single}), training on three independently authored samplers (\textit{indep}), and training on the iteratively refined samplers used by our default method (\textit{iter}).
}
\label{fig:abl_recon}
\end{figure}

We report results of this experiment in  Table~\ref{tab:app_vpi_net}. 
Complementing these quantitative results, in Figure~\ref{fig:abl_recon}, we show representative reconstructions for these three settings.
For this point cloud to shape program reconstruction task, we report chamfer distance (\textit{CD}) and F-score~\cite{TanksAndTemples}.
As can be observed, the approach we adopt with iterative error-correction outperforms the alternative formulations.
Using the union of three independent \textit{sample\_shape} functions is better than using a single sampling functions, but it is better to generate these extra samplers with feedback sourced from the seed set.

\begin{table}[ht!]
    \centering
    \small
    \caption{
    Accuracy of function parameterizations predicted during the application proposal stage (Section~\ref{sec:prop_apps}). We report the average geometric error of directly predicted parameterizations (\textit{Avg Prop Fit}), the best geometric error among directly predicted parameterizations for each target (\textit{Best Prop Fit}), and the best geometric error after searching the expanded parameter set (\textit{Best EPS Fit}). Lower is better for all metrics.
    }
    \begin{tabular}{@{}lccc@{}}
        \toprule
        \textbf{Category} & \textbf{Avg Prop Fit} $\downarrow$ & \textbf{Best Prop Fit} $\downarrow$ & \textbf{Best EPS Fit} $\downarrow$ \\
        \midrule
        Chair   & 0.1098 & 0.0309 & \textbf{0.0244} \\
        Faucet  & 0.1919 & 0.0387 & \textbf{0.0258} \\
        Lamp    & 0.1960 & 0.0491 & \textbf{0.0260} \\
        Table   & 0.2639 & 0.0212 & \textbf{0.0132} \\
        Storage & 0.1505 & 0.0291 & \textbf{0.0275} \\
        \midrule
        Avg. & 0.1824 & 0.0338 & \textbf{0.0234} \\
        \bottomrule
    \end{tabular}
    \label{tab:param_fit_acc}
\end{table}

\subsection{Accuracy of LLM-Predicted Function Applications}
\label{sec:app_res_param_fit}

We also measure how accurately the application proposal stage from Section~\ref{sec:prop_apps} predicts function parameterizations for target geometry. 
In this experiment, \textit{Avg Prop Fit} records the average geometric error of the parameterizations directly predicted by the LLM, \textit{Best Prop Fit} records the best geometric error among the directly predicted parameterizations for each target, and \textit{Best EPS Fit} records the best geometric error after searching the expanded parameter set described in Section~\ref{sec:app_lib_val}.

As shown in Table~\ref{tab:param_fit_acc}, the raw parameter values predicted by the LLM are often only approximately correct, with an average error of 0.1824 across categories. 
However, the best directly predicted proposal for each target is already much more accurate (0.0338 on average), and the expanded parameter search further improves this to 0.0234. 
This analysis helps explain why the proposal stage can still be effective even when direct parameter prediction is not guaranteed to succeed on a per-sample basis: the LLM is often good at identifying the right function and approximate structure, while later search over parameterizations substantially improves the final fit.

\subsection{Varying Seed Set Size}
\label{sec:app_res_seed_size}

\begin{figure*}[t!]
\centering
 \includegraphics[width=\linewidth]{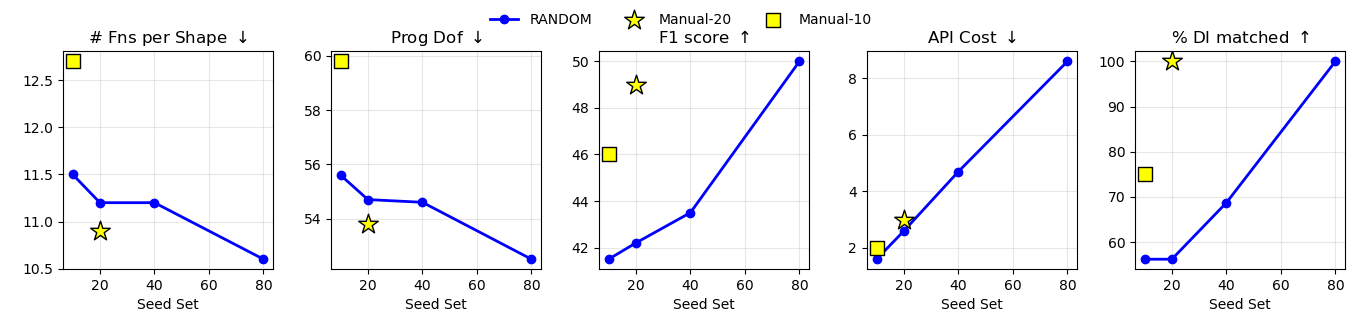}
\caption{
We compare \methodname under random seed sets of varying sizes (10, 20, 40, and 80), extending the seed set variation analysis from Section~\ref{sec:res_method_abl} of the main paper.
In the plot, the blue line denotes randomly selected seed sets, the yellow star denotes our curated 20-shape seed set, and the yellow square denotes a random 10-shape subset sampled from those curated 20 shapes.
We report \# Fns per Shape, Prog Dof, semantic F1 score, API cost, and the percentage of design-intent functions that were successfully matched.
}
\label{fig:seed_set_plot}
\end{figure*}

We further analyze the seed set choice by varying the number of shapes used for library discovery.
For randomly selected seed sets, we consider sizes of 10, 20, 40, and 80.
This extends the seed set variation discussion from Section~\ref{sec:res_method_abl} of the main paper, where we already compared different randomly sampled seed sets of fixed size.
We compare these larger and smaller random settings against our default curated 20-shape seed set, as well as a random 10-shape subset sampled from that curated set.
The blue line in Figure~\ref{fig:seed_set_plot} corresponds to the randomly selected seed sets, the yellow star corresponds to our curated 20-shape seed set, and the yellow square corresponds to the random 10-shape subset from those curated 20 shapes.
The metrics in the plot are \textit{\# Fns per Shape}, \textit{Prog Dof}, semantic F1 score, API cost, and the percentage of design-intent functions that were successfully matched (validated by our procedure).

Figure~\ref{fig:seed_set_plot} shows three main trends.
First, for random seed sets, increasing the seed set size improves performance, especially in semantic alignment and in the fraction of design-intent functions that can be successfully matched.
Second, this improvement comes with a clear increase in API cost.
Third, our curated 20-shape seed set offers a strong trade-off: it validates all design-intent functions, achieves better semantic alignment than smaller random seed sets, and remains much cheaper than the largest random setting.
At a high level, the seed set should be large enough to provide repeated grounded examples of the intended patterns, but carefully chosen examples can achieve a much better quality-cost trade-off than simply scaling random seed sets upward.
This statement also reflects a simple necessary condition for our method: each intended pattern should appear multiple times, since our validation requires a function to be successfully applied across at least two seed shapes.

\subsection{Audit of Design Intent Deviation in Validated Libraries}
\label{sec:app_res_fn_audit}

In Section~\ref{sec:discussion}, we discuss a failure mode where geometrically validated but semantically imperfect functions can still enter the final library.
To better understand this issue, we manually audited all discovered functions used for our main experiments.
For each function, we compared the final validated implementations against the original design intent (function description and the seed-set examples).
In this audit, we found just 6 functions (out of 28) with a mismatch in some part of the final implementation.
These mismatches were generally limited in scope: in most cases, the overall function structure remained faithful to the intended abstraction, but one geometric or parametric detail differed from the stated design intent. 
Below we detail our findings. 
We refer readers to the supplemental material for the referenced library functions and design intents.

\paragraph{\texttt{ladder\_chair\_back} (chair).}

This function reproduces the same failure mode discussed in the main paper.
The design intent specifies a ladder-style backrest whose bars should have thickness-controlled cross-sections.
However, in the discovered implementation, the \texttt{top\_only} and \texttt{top\_bottom} variants assign the full back depth to both the frame and the vertical bars, rather than using a thickness-controlled depth.
In the \texttt{left\_right} variant, the horizontal bars use \texttt{frame\_thickness} for their depth rather than \texttt{bar\_thickness}.
Our audit revealed that this error came about because of a slight mismatch between the function description and the seed-set examples. 
A few of the ladder chair back structures present in the seed set do not have back surface bars with square cross-sections. 
So, lacking the ability to independently set the depth of each bar, the LLM instead fell back to an approximately correct solution of using the full back depth for each bar.

\paragraph{\texttt{cantilever\_base} (chair).}

The design intent for this function includes an optional rear stretcher that should be a square-cross-section bar, parameterized by the stretcher size.
In the discovered implementation, however, the rear stretcher is assigned a fixed depth, rather than using the same size parameter to control both cross-sectional dimensions.
This changes the semantics of the abstraction slightly, even though the resulting geometry remains close to the seed-set examples.
As in the previous case, the mismatch appears to arise because a slightly different parameterization better fits the seed set geometries:
only one of the seed set shapes has a rear stretcher, and in this case the stretcher does not have a square cross-section.

\paragraph{\texttt{pedestal\_base} (table).}

This table function largely matches the intended high-level structure, but its optional top surface is stacked incorrectly relative to the pedestal and central column.
The design intent describes a vertically stacked configuration in which the top surface sits on top of the central support.
In the discovered implementation, that optional top surface is positioned inconsistently with this stacking relationship, even though the overall shape remains plausible.
This appears to be a case where the discovered implementation was specialized to capture a seed-set example whose geometry intertwined two distinct central column parts.

\paragraph{\texttt{wall\_mounted\_lamp} (lamp).}

This example reflects a description-to-interface mismatch rather than a major structural implementation error.
The original design intent specifies that the wall mount should have a square cross-section, so a separate \texttt{wall\_mount\_depth} parameter is not strictly necessary.
However, the discovered function interface introduces this additional free parameter.
The resulting function is still structurally reasonable and remains close to the intended abstraction, but it is less semantically faithful because it exposes a degree of freedom that the original design intent did not intend to vary independently.

\paragraph{\texttt{box\_frame\_base} (storage).}

This is the clearest implementation-side mismatch we found in the audit.
The design intent describes a base made from four frames: two horizontal walls and two lateral walls that together form a supporting base frame beneath the storage body.
The discovered implementation instead uses thickness to control the height of the laterally running side pieces for part of this structure.
While this still produces reasonable geometric fits, as the heights of these box frames are typically small for this category of shape, this modification changes the semantics of the abstraction.
Unlike the other cases, this is not merely a subtle ambiguity in the description; both the seed-set geometry and the function description support a different implementation more strongly than the one that was ultimately discovered.

\paragraph{\texttt{lever\_handle\_set} (faucet).}

The design intent for this function specifies a lever geometry in which, for example, the left end of the right handle should lie over the center of the right support, so that the relative positioning matches the intended pivot geometry.
The discovered implementation uses a slightly different offset rule for the supports and handles.
This produces a visually similar and geometrically acceptable handle arrangement, but not the exact intended parameterization.
Here again, the mismatch appears to come from selecting a parameterization that better reconstructs the seed-set examples under the geometric metric, even though it departs slightly from the stated function description.

Overall, these audited failure modes suggest that while this deviation from design intent can occur, it does not often lead to significant semantic errors in the final library.
Most of the mismatches are subtle and arise because the geometric validation stage can accept functions that reconstruct the seed set well while relaxing a small semantic detail from the original design intent.
Among the six cases, one is primarily a description-to-interface issue that introduces an extra free parameter (\texttt{wall\_mounted\_lamp}), one is a clear implementation-side error (\texttt{box\_frame\_base}), and the remaining cases reflect subtle mismatches in the compatibility of the dual-modes of the design-intent.

\section{Additional Method Details}
\label{sec:app_method}

In this section we provide additional details on the various parts of our method and implementation.

\subsection{Library Design}
\label{sec:app_lib_design}

We provide additional details on the library design logic of \methodname.
Our prompting logic has some shared commonalities -- we begin with a preamble, that tasks the LLM with acting as an expert procedural modeler, that is concerned about structured shape representations.

In our prompts, we define a few helper classes as mentioned in Section~\ref{sec:lib_interface}.
The \textit{Part} class defines an axis-aligned cuboid, from six input parameters: width, height, depth, x position, y position, z position.
This cuboid acts as an part proxy that abstracts out detailed geometry.
The \textit{CoordFrame} class defines a local coordinate frame bounding volume.
Each abstraction function \methodname produces takes in a \textit{CoordFrame} object as its first argument, which controls the extents of the parts produced by the function.
Otherwise, we allow the LLM to use general python syntax.
We find it useful to provide the LLM with a set of general hints for how to complete this task: 
these include, for example, explanations for how \textit{Part} and \textit{CoordFrame} objects should be used, and also explain our conventions for 3D space (width runs along the x axis, height runs along the y axis, depth runs along the z axis).

\subsubsection{Interface creation}
\label{sec:app_interface_creation}

In the interface creation step we use \textit{o1mini} as the LLM.
We ask the LLM to convert each function description into a structured programmatic interface: python function, with typed parameters, and a structured doc-string.
We provide the LLM with a few in-context expert completion examples from held out categories.

We prompt the LLM such that this interface has a particular design.
Each parameter is guided to be an allowed parameter type (float, strings, Booleans, integers).
Each function must return a list of \textit{Part} objects, and must take in a first argument \textit{CoordFrame}.
In the doc-string, there are certain fields we check for.
The `Description' field provides an overview of how the function is expected to be implemented. 
The `Parts' field expresses the possible structural configurations the function can produced. 
The `Parameters' field describes what effect each input parameter has on the output part structure. 
Note that the `Parts' field always need to include a `valid options' list, that states what numbers of parts could be created by the function.
We use this `valid options' list to help prune out obviously bad LLM applications in later stages.
Note that each entry in `valid options' must be greater than 1, as there is no compressive advantage in having an abstraction function that takes in a \textit{CoordFrame} object and returns a single \textit{Part} (that would have to match the extents).
Note we also ensure that each `string' type argument has a defined valid options list in the doc-string, so in practice we treat these `strings' as lightweight Enums.

\subsubsection{Propose Function Applications}
\label{sec:app_prop_apps}

In this stage, we present the LLM with a seed set shape, along with the library interface, and ask the LLM to write a program that would reconstruct the shape.
We represent the shape in a structured fashion, as a list of \textit{Part objects}, where each part has  
dimensions, position, and semantic label.
We also pass a render of the mesh to a VLM (gpt4o), asking the VLM to describe the parts that are present in the shape.

From these two inputs, the task is then to write a program that uses functions to recreate the input part list.
For this purpose, we give the LLM access to a specially designed \textit{group\_parts} functions.
This \textit{group\_parts} function, takes in a list of \textit{Part} objects, and automatically returns a \textit{CoordFrame} bounding object.
We provide the LLM with some out-of-category in-context examples that show how to perform this task for a toy library.
We ask it to try to explain all parts in the shape with calls to the library functions, though if there are any single parts that cannot be explained through library calls we allow the LLM to identify these with independent \textit{group\_parts} calls.

After the LLM predicts a program, we evaluate the function applications.
For each function, we look at the parts that the function is trying to explain (using the \textit{group\_parts}), and can reject bad groupings of parts that are not possible under the constraints of the `valid options' part of the interface.
We record triplets of function name, function parameters, and input parts (identified through the \textit{group\_parts} call).

For each shape, we perform this procedure 5 times. 
The first time we use \textit{o1mini} as the LLM.
The subsequent times we use \textit{gpt4o}, as we find that even reasoning LLM modules struggle to exactly predict correct parameters, so we use the LLM to instead generate approximately correct solutions (that we later validate geometrically).

\subsubsection{Proposing Function Implementation}
\label{sec:app_prop_impls}

In this step we use \textit{o1mini} as the LLM.
We provide the LLM with input-output examples, sourced from previous step, and ask it to propose a python implementation of the function.
Besides the typical hints, we task the LLM with making sure that each parameter, when changed, updates the output geometry (no parameter should be a `no-op').
Beyond this, parts should not float in space (they should have a connected path to the ground).

When we format the input-output examples, we omit all of the parameter values with `?'; we ablate this design decision in Section ~\ref{sec:app_res_alternative_abl}.
From all of the function applications, we select a 
maximum of four in context examples, choosing randomly from the LLM applications, making sure to evenly sample from each shape in the seed set as much as possible.

After the LLM authors a new implementation, we hypothesize that it might find improved parameterizations of the function for the input-output examples 
(now that it knows how the function works).
To leverage this observation, we add an additional request to the LLM: "after you have written your implementation, provide a parameterization for each input example that would recreate its associated output." We then record these new predictions as input-output examples for the validation stage.

\subsubsection{Library Validation}
\label{sec:app_lib_val}

Library validation happens on a per function basis.
For each function, we find every single group of parts that the LLM ever said could be explained by the function in question.
Then we also consider all of the parameterizations the LLM ever suggested for the function.
LLMs are good at generating hypothesis that are approximately correct, and we observe they perform better when the task is more tied to semantics; 
for instance, they do a good job at identifying which functions to use, and which parts they should explain parts, but they are not as adept at exactly predicting continuous parameter values.

With this in mind, we would like to take the set of proposed function parameterizations, and expand it to search for more ways that the function could be executed, which is important in order to check which of the proposed implementations have potential outputs that could validate with respect to structures in the seed set.
So, for each parameter, independently, we identify all unique values observed in the function applications. 
For boolean and string parameters, where we know the possible values ahead of time, and we consider all of these values. 
For floats, we greedily build up a set, with a minimum gap of 0.025, where if we have already added a parameter with value a, and a new parameter comes up with value b, and the distance between a and b is less than this gap, we don't add b into this unique set.
From this unique set, we then consider all of the possible combinations, doing a cartesian product over this list of lists. 
If the size of this product is greater than a max value (10000), then we sub-sample 10000 options from this list.
We always consider the LLM predicted input-output examples first, and then augment this set with the new parameterizations.
We ablate this design decision in Section~\ref{sec:app_res_lib_design}.

Then, for each parameter group produced from this set, we try executing each function implementation.
Against every identified group of parts from the input-output examples, we compute the error between the function output and the goal layout -- 
if this error is below a set threshold, we record a success (see Section~\ref{sec:app_prim_error}).
An implementation needs at least 2 successful validations (on separate seed set shapes) for us to include it into the library. 
We only include up to 1 implementation per function, so we choose the implementation that achieves the lowest error, for the most seed set shapes.

\paragraph{Finding programs for seed set shapes} 
After the validation step we have implemented functions, and we know which function applications can explain which parts in which shapes, but we don't know what is the best program to explain each shape in the seed set. 
We would like to optimize the objective function (see Section~\ref{sec:app_lib_learn_obj}).
To accomplish this, we take a greedy approach, for each (function,  parameterization, and part group) triplet that was validated, we record the objective function cost incurred, averaged over the number of parts the function recreates. 
We then sort the validated function application according to this score, and build up a program by taking the next best function application in a greedy fashion. 
In this step, we make sure to skip function applications that try to recreate parts that a previously used function invocation has already covered.
Any leftover parts are represented with a special \textit{make\_part} function that just defines a \textit{Part} that has a semantic name.

\subsubsection{Primitive Error Function}
\label{sec:app_prim_error}

The geometric error function we use takes in two sets of unordered primitives. 
For every pair of primitives from the predicted to target set, we calculate the maximum minimum distance between any two corners from one primitive to the other. 
We then use a matching algorithm to assign a stable pairing between the two sets.
If any of the distances in this best matching are above a threshold (0.25, where shapes are normalized to lie within the unit sphere), then we say that there is infinite geometric error (the match is invalid).
Otherwise, the geometric error is an average of the MMCD (maximum minimum corner distance) calculated according to the best match.

\begin{figure}[h]
\centering
 \includegraphics[width=\linewidth]{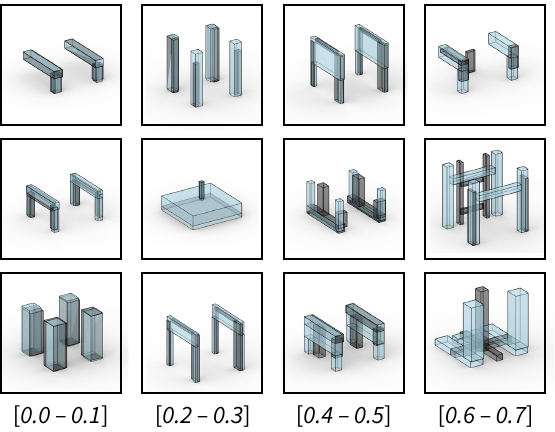}
\caption{
Primitive-set matches at different geometric error ranges under our corner distance-based error metric (Section~\ref{sec:app_prim_error}).
For each bucket, we show representative proposed program snippets (blue) against target geometry (gray), grouped by the maximum geometric error between matched primitives.
}
\label{fig:fit_thresh}
\end{figure}

To provide intuition for this threshold, in Figure~\ref{fig:fit_thresh} we show representative primitive-set matches at different geometric error ranges, with proposed program snippets shown in blue and target geometry shown in grey.
These examples suggest that in the lower-error buckets, especially between 0--0.1 and 0.2--0.3, the intended structure is still fairly well preserved, whereas at larger error ranges the mismatches become much more visually apparent.
This qualitative comparison helps motivate our use of a maximum threshold of 0.25 as a hard validity filter.
This threshold is only used to decide whether a match is valid at all; once a proposal is deemed valid, lower-error matches are still preferred by the library learning objective described below.

\subsubsection{Library Learning Objective Function}
\label{sec:app_lib_learn_obj}

When searching for programs that explain shapes, we need an objective function to guide the search. 
We take inspiration from prior approaches such as ~\cite{jones2023shapecoder}, and formulated an objective function as a weighted average of two terms.
One of these terms counts up the number of degrees of freedom in the program representation, for simplicity we treat every token in the program as a degree of freedom with the same weight (1.).
Another term ensures that the produced geometry does not deviate too far from the target structure. 
We calculate the geometric error (above), and add that into our objective function with a weight of 10.

\subsection{Recognition Networks}
\label{sec:app_rec_net}

We implement all of our networks in PyTorch~\cite{paszke2017automatic}. 
All of our experiments are run on NVIDIA GeForce RTX 3090 graphic cards with 24GB of VRAM.
We use the Adam optimizer~\cite{Kingma2014AdamAM} with a learning rate of 1e-4, dropout of 0.1, and a batch size of 128.
All of our networks train on synthetic data that is sampled `on-the-fly' from the LLM authored \textit{sample\_shape} function, we use 25 million programs for structured input networks, and
16 million programs for unstructured input networks, observing that the loss curves reach saturation at this point.
This training takes between 12 hours and 48 hours, depending on category and input modality.

\subsubsection{Network Architecture}
\label{sec:app_net_arch}

We implement our recognition network as a Transformer decoder. 
Our network has 4 layers, 4 heads, model dim of 256, and a full feature dim of 1024.
This network has full attention over the conditioning information, each visual input is tokenized in a particular fashion (see below).
Programs are similarly tokenized, and our network is trained through teacher forcing, because as we have paired data, we can use cross entropy loss.
This is the same loss that LLMs use in pretraining, or supervised finetuning stages.
We use learned positional encodings, these cap the maximum sequence lengths.

\subsubsection{Structured Inputs}
\label{sec:app_struct_input}

We describe the process for mapping a collection of cuboids to a shape program.
First, each parameter in each primitive in the input shape is quantized and treated as a discrete token.
We order the primitives according to their x-y-z positions, as we do not know how they should be ordered otherwise.
We learn positional encoding for the positions of the input parameters, so our network can reason over up to 20 primitives and programs of up to length 128 tokens.
For primitives, we find it useful to perturb the cuboid parameters of the input a bit, so the network does not overly pick up on `perfect' patterns produced by executing the cuboids.
To this end, we add a noise of gaussian level 0.05 to each cuboid parameter.
Note that this changes the input to the network, but the target program does not change, this acts similar to dropout, as a kind of regularization on learning.

\paragraph{Inference procedure}
Our inference procedure prompts the network with an input set of unordered primitives and samples a large number of programs according the network's predicted distribution.
We try executing each program, and we record its complexity (the number of tokens it uses) and its geometric error against the input set (using the a geometric error metric, Section~\ref{sec:app_prim_error}). 
We sample programs independently up until a timeout (4 seconds per shape) -- note this is half the time that ShapeCoder uses on average per shape inference.
We choose the program that minimizes an objective that is a simple weighted combination of these two values, see Section~\ref{sec:app_lib_learn_obj}.

As we know the ground truth set of primitives we are trying to get the program to recreate, we can also integrate a more advanced search strategy.
Say we have one version of the program $a$, that has achieved the best objective value so far.
Now say we predict another program version $b$, that achieves a worse objective value.
While we could just throw the entire prediction of $b$ away, instead we can consider all of the sub function applications made in $b$, and check if any of these calls, if merged into the program version $a$, would improve its objective function score.
This more involved search strategy offers only a small, but significant improvement, and mirrors similar primitive based reconstruction approaches adopted by prior work~\cite{jones2023shapecoder}.

\subsubsection{Unstructured Inputs}
\label{sec:app_unstruct_input}

We describe the process for mapping unstructured geometry (e.g. a point cloud or an occupancy voxel field) to a shape program.
For point cloud inputs, we replace the primitive token encodings with an embedding produced by a PointNet++~\cite{qi2017pointnet++} network.
This network produces 4 visual tokens.
For voxel inputs, we replace the primitive token encodings with an embedding produced by a 3D-CNN.
This network produces 8 visual tokens.

To convert a program into a point cloud, we first execute the program to produce a collection of cuboids.
We convert these cuboids into a mesh (not worrying about intersections), and then sample a point cloud from the surface of the mesh (2048 points for all experiments, without normals).
To convert a program into voxels, we take a set of query points (voxel centers), and evaluate the local position of these query points within each cuboid.
If the query point is inside any cuboid, or the distance to the nearest cuboid is less than half the length of a voxel, we say the voxel is `occupied', otherwise the voxel empty. 
We use voxel fields of size 64x64x64.

\paragraph{Inference procedure}
During this inference procedure, unlike the structured visual program induction setting, we don't know which parts should be created.
So here, we just sample a large number of programs from the network: 1000 independent samples, using Top-P sampling (with P=90\%). 
For each sampled program, we try executing it, convert the output cuboids into a point cloud, and record the reconstruction performance.
We choose point cloud reconstructions based on the Chamfer distance between the input and sampled mesh.
We choose voxel reconstructions based on the IoU between input voxels and output voxel field.
F-score~\cite{TanksAndTemples} is another metric we consider for point cloud reconstructions, using a threshold of 0.03 (where meshes are normalized to lie within unit bounding box).

\subsection{Synthetic Data Sampler}
\label{sec:app_synth_data_sampler}

We task the LLM, provided with the interface and validated seed set programs, to author a \textit{sample\_shape} function.
We tell it that this function should randomly produce new procedural shapes using the library of functions, by taking in a single CoordFrame parameter, that specifies the global bounding volume of the shape to be produced. 

We use LLM models with reasoning capabilities (e.g. o1), 
as this task is quite difficult, to author a convincing data sampler. 
After the LLM authors a version, we see how well it actually matches the patterns in the seed set.
We perform two rounds of automated feedback for each function in the dataset.
In each round of feedback, we evaluate the function by sampling a diverse set of shapes and assessing various aspects of its behavior. 
We examine whether all functions in the library were used, whether all parameter types were employed, and whether all output structures described in the function's documentation were produced. 
These checks are performed automatically.
Additionally, we analyze the structures generated by the sampled function applications to determine their similarity to those observed during the validation stage. 
If any significant deviations are detected, the sampler is instructed to update its logic to produce outputs closer to the expected structures.
The goal of this procedure is not necessarily to produce a `better' sample shape function, but rather to produce variations that explore different parts of the seed set space.
More specifically, this iterative feedback procedure is designed primarily to improve geometric and structural coverage, not to directly enforce semantic correctness.
Its purpose is to encourage the sampler to cover the major modes of variation observed in programs found from the seed set shapes, rather than to detect or repair subtle semantic mismatches in the discovered abstractions.
During training of recognition networks we randomly sample from all LLM produced sample shape functions; we randomly choose a sampling function, then randomly sample a program, to see a more diverse spread of programs.
See Section~\ref{sec:app_res_rec_net} for an experimental ablation on this design decision.

\subsection{Generation and Editing Applications}
\label{sec:app_gen_edit_apps}

\subsubsection{Detailization}
\label{sec:app_detail}

We develop a detailization module, that can convert a layout of cuboids to a 3D mesh with geometric details.
We take a pretrained text-to-3D shape generative model, Cube~\cite{roblox2025cube}, 
and adapt it to instead condition on a layout of primitives (not text). 
We first describe the base Cube model, and then describe our adaptation.

\paragraph{Base Cube Model}
The base cube model is composed of two parts.
One part of the system is an encoder-decoder, the encoder takes in a point cloud and outputs a sequences of discrete codes (512 in length).
The decoder takes this sequence of discrete codes, 
and uses marching cubes to produce a 3D mesh, after predicting per-point occupancy values.
We do not change the encoder or decoder part of the pipeline, keeping them frozen. 
Then a decoder only transformer module learns to model sequences of these discrete codes in an autoregressive fashion.
Given an input shape with an accompanying text prompt, they embed text prompt into a set of 77 tokens using a frozen clip encoder.
The transformer attends over these 77 tokens, and autoregressively predicts next 512 shape tokens one at a time, which can then be converted into a 3D mesh with the decoder.

\paragraph{Structured adaptation}
In our adaptation, we start with the pretrained, released model.
We remove the CLIP conditioning, instead we add in a box encoder.
This box encoder consumes a layout of cuboid primitives, lifts each primitive to a higher dimension with a linear layer, and then we use attention layers to blend this feature space together.
This process produces up to 20 new tokens, which the transformer attends over.
We finetune the attention weights of the transformer, and all parameters of the box encoder, while keeping all of the feed forward layer weights frozen. 

We finetune this network with paired cuboid layouts and meshes with geometric details; sourcing this data from PartNet~\cite{PartNet}, training in a per-category fashion.
We normalize the cuboid layout, so that the entire layout is centered and has extents that match the unit sphere.
After decoding a mesh, we then transform the output shape to match the original dimensions of the input layout.
To formulate training data, for each input mesh, we sample the surface to produce a point cloud, and then encode this point cloud into a sequence of discrete codes using the frozen encoder. 
We then condition the network on the features produced by the box encoder, and update the network with cross entropy loss, using teacher forcing, on the \textit{GT} sequence of discrete codes.
We train on between 1000 and 6000 shapes per category, which takes between 2-4 days until loss convergence, with a batch size of 4, a learning rate of 1e-5, with 24GB of ram.

To integrate this network with \methodname programs, we take programs, execute them to get a layout of cuboids, feed this layout into the network, and get out a sequence of discrete codes.
This sequence of discrete codes can be presented to the frozen decoder to produce a 3D mesh.
We sample from the transformer without classifier free guidance, with Top-P sampling of 90\%, so we get variations in the style, and quality, of the detailizations.
In future work, it would be interesting to try to automatically identify which samples have a better (or worse) adherence to the input part layout conditioning.
We visualize some of the detailizations in Figure~\ref{fig:geom_gen}.

\subsubsection{Structured Deformation}
\label{sec:app_struct_deform}

In Figure~\ref{fig:edit_geom}, we demonstrate that \methodname program edits can drive mesh edits through a deformation scheme. 
We use a simple deformation scheme inspired by cage-based deformation schemes. 
Assume we have:
a starting mesh, with vertices and faces;
a starting cuboid layout, produced by program version 1;
an edited cuboid layout, produced by program version 2.
Critically, we assume this layout change was made by a program edit, so we know the correspondence between the cuboids in the starting layout and the cuboids in the edited layout.

With this input, our scheme operates as follows.
For each mesh vertex, find the local position of the vertex within each cuboid in the starting layout.
Then assign a weight of this vertex to each cuboid.
How do we assign this weight?
If the vertex is inside a single cuboid, all of its weight goes to that cuboid.
Otherwise, if the vertex is inside multiple cuboids, or no cuboids, we assign the weight as distribution over all of the cuboids.
This distribution is determined according to a soft blend, that uses the inverse of the distance of each point to the surface of each cuboid to determine the shape of the distribution.

With this weighting scheme, given a new primitive layout, we can produce new vertex positions.
We calculate the local position of each vertex with respect to each new cuboid placement, and then average these positions with respect to the cuboid weighting distribution.
Note, that in this scheme, the face structure of the mesh does not change.

Our formulation does not currently support removing or adding parts, so we assume that program edits change only positions and or the dimensions of the cuboids.
This editing scheme could be replaced with other, more sophisticated solutions (\cite{Uday_deform}),
or be improved by adding in regularization terms (\cite{ARAP_modeling:2007})
but as demonstrated in Figure~\ref{fig:edit_geom}, this simple approach already works well enough to produce compelling sets of edits.

\section{Additional Experiment Details }
\label{sec:app_exp}

In this section we provide additional details on our experimental design. As an overarching note, we point out that in the figures of all rendered program executions, we maintain a consistent color for the cuboids created by a certain function;
e.g. every cuboid created by a `ladder chair back' call is given a green color (see Figure~\ref{fig:prog_comp}). 
When cuboids are created without a semantic abstraction (e.g. a Part is directly instantiated) we color the output parts in gray.

\subsection{Cost and Timing}
\label{sec:exp_cost_time}

We provide detailed estimates for how expensive it is (from a time and API monetary expense perspective) to use our system to discover libraries of shape abstraction functions.
To produce 20 shape descriptions from images using \textit{gpt4o}: 10 cents and 1-2 minutes.
To create library interfaces from textual descriptions with \textit{o1mini}: 25 cents, 2-4 minutes.
To propose function applications over (20) shapes with (1) \textit{o1mini} call and (4) \textit{gpt4o} calls: \$2-3 
and 15-25 minutes.
To propose (4) implementations for each function with \textit{o1mini}: \$2-4 and 15-30 minutes.
To propose a single program sampler with \textit{o1}: 50 cents and 1 minute.
Validation takes between 5 and 20 minutes per category, depending on the number of functions and the size of the parameter sets to evaluate.

\subsection{Data}
\label{sec:exp_data}

Collections of example shapes in the seed set are chosen by an user, with the idea that they should match the design intent expressed in the function descriptions.
The design of the function descriptions was done up-front, without iteration, where a expert expressed domain knowledge to the LLM in a easily transmissible medium.
See Section~\ref{sec:app_res_alternative_abl} where we ablate this user input.
Specifically, we have the user select 20 partNet shapes and put them in a list, and then we can automatically produce the rest of the structured data from the PartNet annotations. 
ShapeNet meshes with texture are then rendered, using metadata in PartNet.

After creating seed sets, we created validation sets of held out PartNet shapes: chairs (1000), storage(400), tables (1000), faucet (400), lamps (656). 
All experimental results are done on the validation shapes, never seen before by \methodname.

\subsection{ShapeCoder}
\label{sec:exp_shapecoder}

\paragraph{Library Generation}

ShapeCoder is the best existing method for library learning over 3D shapes.
ShapeCoder tries to discover abstraction functions that improve a compression-based objective over a dataset.
In our comparisons against ShapeCoder we use the officially released implementation, and run it over the seed sets.
The only change we make is removing the rotation operation from the base ShapeCoder language, as we focus on structures of axis-aligned primitives in our experiments.

\paragraph{Recognition Network Training}

\methodname uses a LLM produced sample shape function to get training data for its recognition network.
ShapeCoder takes a different approach, sampling random parameterizations of functions, and then creating synthetic data by randomly combining these functions into random scenes.
This scheme was original proposed \textit{only} to train recognition networks (with a related architecture) that map cuboids to shapeCoder programs. 
For our structured reconstruction tasks (function usage and semantic consistency in Table~\ref{tab:lib_learn_comp}) we use these networks.
To allow ShapeCoder to perform unstructured visual program induction tasks (mapping point clouds or voxels to shape programs), we adapt this framework, by taking our recognition networks and training them to predict ShapeCoder programs. 
For these networks, we follow the original ShapeCoder formulation using random combinations of functions to produce the training scenes, as ShapeCoder cannot leverage the prior from a LLM to produce our \textit{sample\_shape} functions.

\paragraph{LLM Usage Experiments}

We experiment with how well LLMs can interact with ShapeCoder programs, and find that this is challenging for frontier models.
ShapeCoder functions and parameters do not have semantic-aligned names, and tend to over-focus on local patterns, especially when discovered from small seed sets.
To try to help the LLM as much as possible, we provide prompts where each abstraction is presented with its definitions, along with an expert documented base ShapeCoder library.

\subsection{\llmbaseline}
\label{sec:exp_llm_only}

\paragraph{Library Generation}

\llmbaseline is an ablated version of our method that relies on only the prior of the LLM and the design intent of the expert user in the form of function descriptions.
We compare \methodname against this condition to validate the need for using the seed set of shapes alongside the natural language specification.
By default we use \textit{o1mini} as the LLM for this condition, but experiment with other LLM versions in Section~\ref{sec:app_res_alternative_abl}. 

This baseline is equivalent to \methodname modulo a few critical changes. 
The interface creation step is exactly the same. 
After this step though, \llmbaseline immediately implements each function, without using any input/output guidance about how this function should constructed. 
As it has no seed set, it assumes that the LLM has perfectly implemented each function, and cannot validate whether its production was `good', or just a hallucination.

\paragraph{Recognition Network Training}

After the LLM has produced a library of shape abstraction functions, it next advances to the synthetic sampler design stage where it prompts the LLM to produce a \textit{sample\_shape} function.
However, as \llmbaseline doesn't have access to a seed set, it cannot source in-context examples of how functions can be used to represent seed set shapes.
This also means that the \textit{sample\_shape} function cannot be improved with our iterative error-correction approach.
Once the LLM has generated the \textit{sample\_shape} function,  we can follow the logic of~\methodname, training a recognition network on data produced by randomly sampling this procedure.

\paragraph{LLM Usage Experiments}

Like \methodname, \llmbaseline has a function library that exposes an interpretable programmatic interface.
At the same time, its functions don't produce outputs that are as useful in representing `real' shape geometries, as we've demonstrated extensively.
In terms of how this impacts LLM interaction, as \llmbaseline doesn't have access to the seed set, for various tasks (like the text-to-shape program generation), the baseline does not get in-context examples of text and program paired data, whereas ShapeCoder and \methodname can use the seed set to source this information.

\subsection{LLM Shape Editing}
\label{sec:exp_llm_shape_edit}

In Section~\ref{sec:res_edit}, we presented an LLM editing experiment. 
Here, we provide additional details on the design of this experiment.
We sourced 5 shape programs from the validation set of each category, and consider 4 edits per shape, giving us a cross-product of 100 total comparison conditions.
We use \textit{o1mini} as the LLM, and observe that \textit{o1mini} produced ShapeCoder programs that had a failed execution for 11/100 conditions (so we omit those from the study).
Note, that none of the edited \methodname programs had execution errors.

We recruited 13 university students to perform the perceptual study. 
Each participant made 25 judgments on two questions for each shown comparison. 
Question 1: "Which edited shape better matches the intent of the request?".
Question 2: "Which edited shape has a more plausible structure?"

\subsection{LLM Shape Program Generation}
\label{sec:exp_llm_shape_prog_gen}

We are interested in measuring the usability of the libraries of programmatic shape abstraction functions discovered by different methods.
To this end, we designed an LLM shape program generation experiment, see Section~\ref{sec:res_gen}.
We wanted to see if an LLM (taking the place of a user interacting with shape programs) could use the library to create reasonable novel shape structures.
To get diversity in the outputs of the LLM, we give it a text description of some shape that the program should exemplify. 
Note, that we do not directly measure how well the output structure adheres to the input prompt, but rather we use the set of text prompts to generate a diverse set of structures.
We can then measure how close this set of structures is to a set of GT shape structures sourced from validation shapes of the same category.
That said, as evidenced by the qualitative results in Figure~\ref{fig:prog_gen}, beyond producing more plausible shapes, \methodname also does a solid job of adhering to the semantic request of the input -- though we found it difficult to directly measure this property.

How do we generate the set of prompts in this experiment?
For each category, a user annotates simple descriptions of 10 seed set shapes.
Then we use these as a seed for LLM ideation, asking the LLM, iteratively, to take these text prompts as a starting point and come up with new descriptions in the same style.
We repeat this 10 times, producing 5 new descriptions each prompt, to get 50 total text descriptions per category.

\subsection{VLMs for visual program induction}
\label{sec:exp_vlm_details}

As discussed in Section~\ref{sec:res_recon_variations}, we evaluated replacing our recognition networks with off-the-shelf VLMs for the task of mapping input shapes to programs that use abstraction functions.
We used gpt-4o as our VLM (i.e. a LLM with vision capabilities).
We prompted the VLM with a (i) a description of the task, (ii) information about the library functions discovered by \methodname, and (iii) 3 examples of seed set shapes represented with \methodname programs.
For (ii), for each function we include the full implementation and doc-string, in the prompt.
For (iii), as current VLMs cannot natively handle 3D data (point clouds / voxels) we evaluated two shape representations: primitives (structured shapes) and images (unstructured shapes). 
For primitives, we actually do not need vision capabilities, as we can pass in the primitive representation in text (e.g. as a nicely-formatted list of the primitive attributes).
For images, we first render the mesh using any texture information from ShapeNet~\cite{chang2015shapenet} when available.
Then, we pass in a composite image with four quadrants, where each quadrant has a render of a separate shape. 
The shape in the top-left corner corresponds to the first in-context example.
The shape in the top-right corner corresponds to the second in-context example.
The shape in the bottom-left corner corresponds to the third in-context example.
We then put the new validation shape in the bottom-right corner, and ask the VLM to predict a program that would reconstruct this new example.
In our experiments, we run this procedure over 40 chair shapes from the validation set.
The VLM from both structured or unstructured inputs performs much worse compared with our library-specific recognition networks that take in unstructured 3D-native representations.

\section{Prompt Templates}
\label{sec:app_prompts}

We include prompt templates used within \methodname. 

\subsection{System Prompt}
\label{sec:app_prompt_system}

This shared prompt-snippet establishes the overall role of the model.

\begin{lstlisting}
You are an expert in designing procedural models of 3D shapes. You want to find programmatic representations of 3D shapes that expose meaningful degrees of freedom. The procedural models you design should allow users to easily manipulate and interact with 3D models. These procedural representations capture the structure of parts that make up 3D shapes, they are not concerned about texture or material properties.
\end{lstlisting}

\subsection{Default Operations}
\label{sec:app_prompt_default_ops}

This shared code context defines the primitive building blocks that appear throughout the prompting pipeline. In particular, it gives the model a consistent representation for part proxies and local coordinate frames before it is asked to propose interfaces, programs, or implementations.

\begin{lstlisting}
from typing import List

class Part:
    def __init__(
        self, width: float, height: float, depth: float, x_pos: float, y_pos: float, z_pos: float
    ):
        """
        Creates part proxy cuboid with dimensions (width, height, depth) and position (x_pos, y_pos, z_pos). This cuboid is always axis-aligned, no orientation or rotation needs to be specified.
        """
        self.dimensions = (width, height, depth)
        self.position = (x_pos, y_pos, z_pos)
    
class CoordFrame:
    def __init__(
        self, width: float, height: float, depth: float, x_pos: float, y_pos: float, z_pos: float
    ):
        """ 
        Defines a local coordinate frame. This coordinate frame is axis-aligned so it is specified with dimensions (width, height, depth) along with a center position (x_pos, y_pos, z_pos).
        """

        # Dimensions
        self.width = width
        self.height = height
        self.depth = depth
        # Position
        self.x_pos = x_pos
        self.y_pos = y_pos
        self.z_pos = z_pos
\end{lstlisting}

\subsection{Describe Shape}
\label{sec:app_prompt_describe_shape}

This prompt is used when we ask a VLM to summarize the structure present in an input example.

\begin{lstlisting}
You are a helpful assistant that has expertise in procedural modeling and shape analysis.

You will describe how shapes decompose into their constituent parts.
{EXAMPLES}

Provide a structural description of the object in the input image. Write your description in the same style as the above descriptions, in a single paragraph, and do not mention texture or material.
\end{lstlisting}

\subsection{Make Interface}
\label{sec:app_prompt_make_interface}

This prompt converts a natural language function description into a library abstraction interface (Section~\ref{sec:lib_interface})

\begin{lstlisting}
{SYSTEM_PROMPT}

You are helping to develop a domain-specific library that contains functions that are helpful for modeling {CATEGORY} objects. Your goal is to improve this domain-specific library by defining new functions that capture common patterns. These patterns should capture structural patterns, e.g. how functions and discrete variables are related and parametric patterns, e.g. how continuous variables are related. Your goal is to produce parsimonious procedural models, so the functions you write should not use overly many parameters; they should use as few parameters as possible, without limiting the expressivity of the function. Moreover, each function should be self-contained: it should not reference any other shape parts beyond those in its input arguments.

You are provided with the following classes. The Part class creates part proxies that you will use to model shape structures. The CoordFrame class defines a local coordinate frame, and will be consumed by the abstraction functions you write.

```python
{DEFAULT_OPS}
```

An expert user will provide you with a natural language description of a new function they would like to add into the library. Your task is convert this natural language description into a function signature and doc-string. Here are some demonstrations of how an expert user would perform this task.

{EXPERT_DEMONSTRATIONS}

Do not return function implementations, only return the function names, their parameters, and doc-strings. Here are some hints to help you with this task.

Hint 1: Each function should take in a CoordFrame object as its first argument, defining the size and location of the functions output. The structures produced by the implementation of this function must match the CoordFrame bounding box exactly, so it will require at least one part (or a combination of parts) to "stretch" to all edges of that bounding box. Further, the generated parts shouldn't float in the air, they should connect to other parts that are together rooted at the bottom of the CoordFrame. 

Hint 2: Each function argument should be typed as either a CoordFrame, a float, an integer, or a string. If a string argument is used, list all possible valid parameterizations in the doc-string of the function. Arguments should not be made Optional, although they can use default values. Moreover, each input parameter should change the output structure when it is modified. Rephrased, no input parameter should be a 'no-op'. 

Hint 3: The dimensions of some parts may be implicitly specified by a combination of the bounding coordinate frame and other input parameters.

Hint 4: All parts and structures will be explained using the directions and orientations of the bounding CoordFrame. Width will correspond with the x axis, where positive values are to the right and negative values are to the left. Height will corespond with the y axis, where positives values are up and negative values are down. Depth will correspond with the z axis, where positive values are towards the front and negative values are towards the back. In this way, a bar that runs horizontally will have a (height x depth) cross-section, a bar that runs vertically will have a (width x depth) cross-section, and a bar that runs laterally will have a (width x height) cross-section.

Hint 5: Follow the above expert demonstrations to properly format the doc-string. Each doc-string should have three fields. The 'Description' field provides an overview of how the function is expected to be implemented. The 'Parts' field expresses the possible structural configurations the function can produced. The 'Parameters' field describes what effect each input parameter has on the output part structure. Note that the 'Parts' field always need to include a 'Valid options' list.

Use these hints and the above expert demonstrations to guide your reasoning for the following description.

Input: {FN_DESCRIPTION}

Output:
\end{lstlisting}
\subsection{Application Proposal}
\label{sec:app_prompt_app_prop}

This prompt asks the model to reconstruct a target seed-set shape using the current library, even though the implementation is not provided (Section~\ref{sec:prop_apps})

\begin{lstlisting}
{SYSTEM_PROMPT}

You will take in a 3D shape as input, where the 3D shape is represented as a list of Part objects that represent a 3D object. You will also be given access to a domain-specific library that contains helpful functions. You will need to decide how to use the set of functions in the current library to reconstruct the input shape.

We provide an example of how an expert would perform this task. 

{EXPERT_EXAMPLE}

You will perform a similiar task for {CATEGORY} shapes. Use the above example to guide your logic. You are provided with the following library of functions designed to model {CATEGORY} shapes.

```python

{DEFAULT_OPS}

def group_parts(
  parts_in_shape: List[Part], group_indices: List[int]) -> CoordFrame:
  """
  Helper function that creates a CoordFrame object that contains all of the parts from parts_in_shape from the specified indices.
  """

{LIBRARY_INTERFACE}

```

The output you return should be formatted as a block of python code. The python code should define a program() function that uses the above library functions as sub-calls to find a compact program representation of the input example. Don't import any libraries, assume they have already been imported, and do not return any function implementations.

Here are some hints to help guide you for this task:

Hint 1: Not all provided functions in the library need to be used in the program you create, you must identify which of them are relevant to the given input shape.

Hint 2: Each library function takes in a CoordFrame object as its first argument. You should use the group_parts function to automatically generate a CoordFrame that bounds Part objects from the parts_in_shape list.

Hint 3: Each library function output should reproduce the Part objects passed into the group_parts function to produce its CF. Make sure that each library function call you make could output the number of Part objects passed into the group_parts function. Library functions express the possible numbers of Parts they can create in the 'Valid options' part of their 'Parts' field within their doc-string.

Hint 4: It may not be possible to reconstruct all parts of the input shape with the provided functions; in this case, use individual group_parts function calls to handle such parts.

Input:
```python

# {SHAPE_DESCRIPTION}
parts_in_shape = [
  {LIST_OF_PART_TARGETS}
]
```

Output:
\end{lstlisting}

\subsection{Implementation Proposal}
\label{sec:app_prompt_impl_prop}

This prompt asks the LLM to implement the function given its interface together with its example input-output pairs (Section~\ref{sec:prop_impls}).

\begin{lstlisting}
{SYSTEM_PROMPT}

You are provided with the following classes. The Part class creates part proxies that you will use to model shape structures. The CoordFrame class defines a local coordinate frame, and will be consumed by the abstraction functions you write.

```python
{DEFAULT_OPS}
```

Your goal is to implement a function named {FN_NAME}. Here are some hints to help guide you for this task: 

Hint 1: The shapes produced by your implementation must match the CoordFrame bounding box exactly. So you need at least one part (or a combination of parts) to "stretch" to all edges of that bounding box. 

Hint 2: Each input parameter should change the output structure when it is modified. Rephrased, no input parameter should be a 'no-op'. 

Hint 3: The dimensions of some parts may be implicitly specified by a combination of the bounding coordinate frame and other input parameters.

Hint 4: Parts in the created structures shouldn't float in the air, they should connect to other parts that are together rooted at the bottom of the CoordFrame. For instance, when one part should be vertically supporting another part, confirm that the top of the lower part is not below the bottom of the upper part.

Hint 5: All parts and structures will be explained using the directions and orientations of the bounding CoordFrame. Width will correspond with the x axis, where positive values are to the right and negative values are to the left. Height will corespond with the y axis, where positives values are up and negative values are down. Depth will correspond with the z axis, where positive values are towards the front and negative values are towards the back. In this way, a bar that runs horizontally will have a (height x depth) cross-section, a bar that runs vertically will have a (width x depth) cross-section, and a bar that runs laterally will have a (width x height) cross-section.

You have previously come up with the following function signature and description for this function.

```python
{FN FROM LIBRARY_INTERFACE}
```

To guide your implementation, you are given some input and output example pairs that describe the intended behavior of this function. The input examples use a special '?' token to indicate that the correct value for this parameter is not known for the associated output pair. Your implementation should create the output structure of parts when the correct parameters are substituted for the '?' tokens.

Example 1
Input: {FN_NAME}({PARAM_EXAMPLES_1}
Output: [
  {LIST_OF_OUTPUT_PARTS_1}
]

Example 2
Input: {FN_NAME}({PARAM_EXAMPLES_2}
Output: [
  {LIST_OF_OUTPUT_PARTS_2}
]

...

Write an implementation of {FN_NAME} using the above information. Try to ensure that your implementation is consistent with the function signature, doc-string, and examples provided. Return your answer as a block of python code. Do not duplicate the doc-string in the implementation you write.

After you have written your implementation, provide a parameterization for each input example that would recreate its associated output. This should be formatted in the same block of python code as your implementation, where each example parameterization is separated by a comment.
\end{lstlisting}

\subsection{Sampler}
\label{sec:app_prompt_sampler}

This prompt is used after a library has been validated. It asks the model to write a \texttt{sample\_shape} function that composes the learned abstractions into diverse but plausible shapes based on an input global coordinate frame, using example function applications and seed-set programs as guidance. See Section~\ref{sec:lib_usage}.

\begin{lstlisting}
{SYSTEM_PROMPT}

You have previously developed the following library of abstraction functions:

```python
{DEFAULT_OPS}

def make_part(
    part_name: str,
    width: float,
    height: float,
    depth: float,
    x_pos: float,
    y_pos: float,
    z_pos: float
) -> Part:
    """
    Creates a Part object with the specified parameters
    """

{LIBRARY_DEFINITION}
```

Here are some examples of how an expert would parameterize these functions while modeling {CATEGORY} shapes:

{FN_EXAMPLE_USAGES}



Here are some examples of how an expert would use these functions to model instances of {CATEGORY} shapes:


{SEED_SET_PROGRAMS}


Your goal is to author a new function, sample_shape, that randomly produces new procedural {CATEGORY} shapes using the library of functions. Your sample_shape function should take in a single G_CF CoordFrame parameter, that specifies the global bounding volume of the shape to be produced. Your sample_shape function should return a list of Part objects. Each call to sample_shape should return a list of parts that correspond with a {CATEGORY} shape when they are unioned together. Here are some hints to help you with this task:

Usage hint 1: All Part objects created by your implementation should remain inside the bounding volume defined by G_CF. Beyond this, if you combine all of the Part objects returned by your implementation, and compute a new bounding volume by considering their maximum extents, this should create a bounding box equal to G_CF. This means that for each of the outer faces in X, Y, and Z directions of the global coordinate frame, some Part must extend to a portion of this bounding plane. Note that the bounding volume may contain empty space in between parts, and some area on each of the outer faces of G_CF may be uncovered. You can assume that G_CF will not extend beyond -1 in the negative direction or extend beyond 1 in the positive direction along each X, Y, and Z axis.

Usage hint 2: The structures returned by each library function will have a bounding box that exactly matches the extents of the input CF parameter to the function.

Usage hint 3: The dimensions of some parts created by each function call may be implicitly specified by a combination of the bounding coordinate frame and other input parameters.

Usage hint 4: Parts in the created structures shouldn't be disconnected from the main shape structure. Put another way, each part should have a path, through part-to-part connections, to the ground plane of the G_CF.  

Usage hint 5: If you want to create a part not covered by a function in the library, you can use the make_part function.

Usage hint 6: All parts and structures will be explained using the directions and orientations of the bounding CoordFrame. Width will correspond with the X axis, where positive values are to the right and negative values are to the left. Height will correspond with the Y axis, where positives values are up and negative values are down. Depth will correspond with the Z axis, where positive values are towards the front and negative values are towards the back. In this way, a bar that runs horizontally will have a (height x depth) cross-section, a bar that runs vertically will have a (width x depth) cross-section, and a bar that runs laterally will have a (width x height) cross-section.

Use the above examples as inspiration for the implementation you develop. Try to make sure that the kinds of part structures observed in these examples could be generated by your sample_shape function through random sampling.
\end{lstlisting}

\subsection{Sample Iteration}
\label{sec:app_prompt_sample_iter}

This follow-up prompt is used to refine the sampler after automated analysis reveals poor coverage of certain seed-set patterns. Rather than asking the model to memorize specific examples, it asks for an updated sampling strategy that better covers the missing structural modes.

\begin{lstlisting}
Your implementation was evaluated by an expert who identified some areas for improvement in its logic. The expert annotated example usages of each library function, and then identified cases when your random sampling logic did not produce parameterizations close to the expected values even after many random samples.

Here is the feedback from the expert:

{SEED_SET_BAD_COVERAGE}

Please update your implementation with this feedback in mind. Try to improve it so that these kinds of examples could be produced through random sampling. Do not add logic that would exactly replicate these specific examples, but instead try to reason about why your previous implementation was not able to produce any similar outputs, in order to find an improved sample_shape function.
\end{lstlisting}

\section{Example Design Intent}
\label{sec:app_design_intent}

We include representative examples of function descriptions provided by \methodname users as part of their design intent input. 
For each category, we include a function description and renders of seed set shapes that contain the described structural pattern.

\textbf{Chair example:}

\begin{lstlisting}
Add a sled_base function that can be used to create a base for chair objects. It should include four vertical legs positioned at the corners of the bounding volume and two lateral sled runners at ground level. The sled runners should connect pairs of legs (front and back on each side). The vertical legs should extend from the runners up to the seat, with their height determined by the bounding volume's height. One parameter should control the cross-sectional size of the legs, and the sled runners should use matching dimensions for their width. Another parameter should control the height of the sled runners. This function should always return 6 parts: 4 legs and 2 sled runners.
\end{lstlisting}

\begin{center}
\includegraphics[width=0.32\columnwidth]{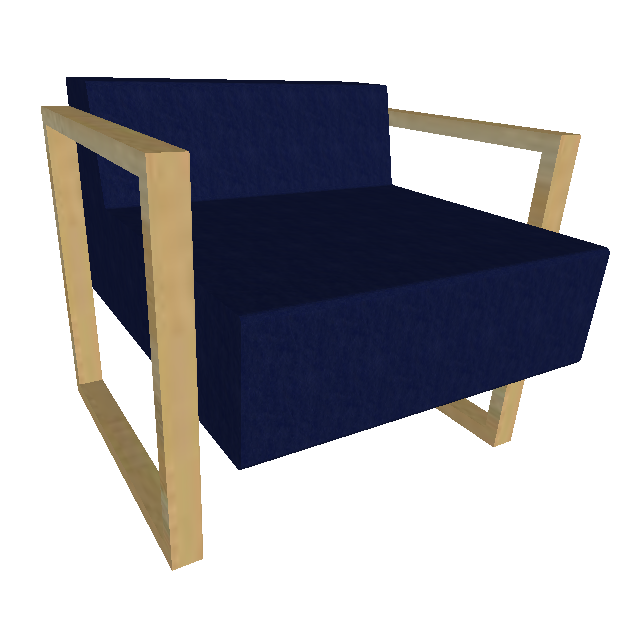}
\hfill
\includegraphics[width=0.32\columnwidth]{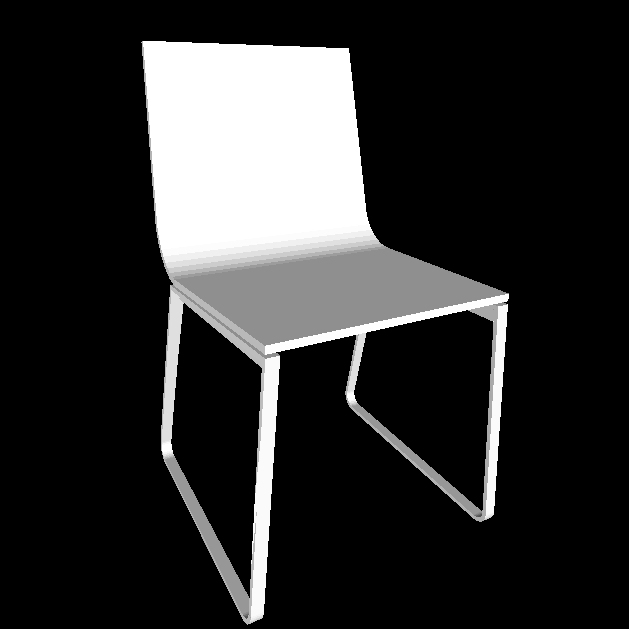}
\hfill
\includegraphics[width=0.32\columnwidth]{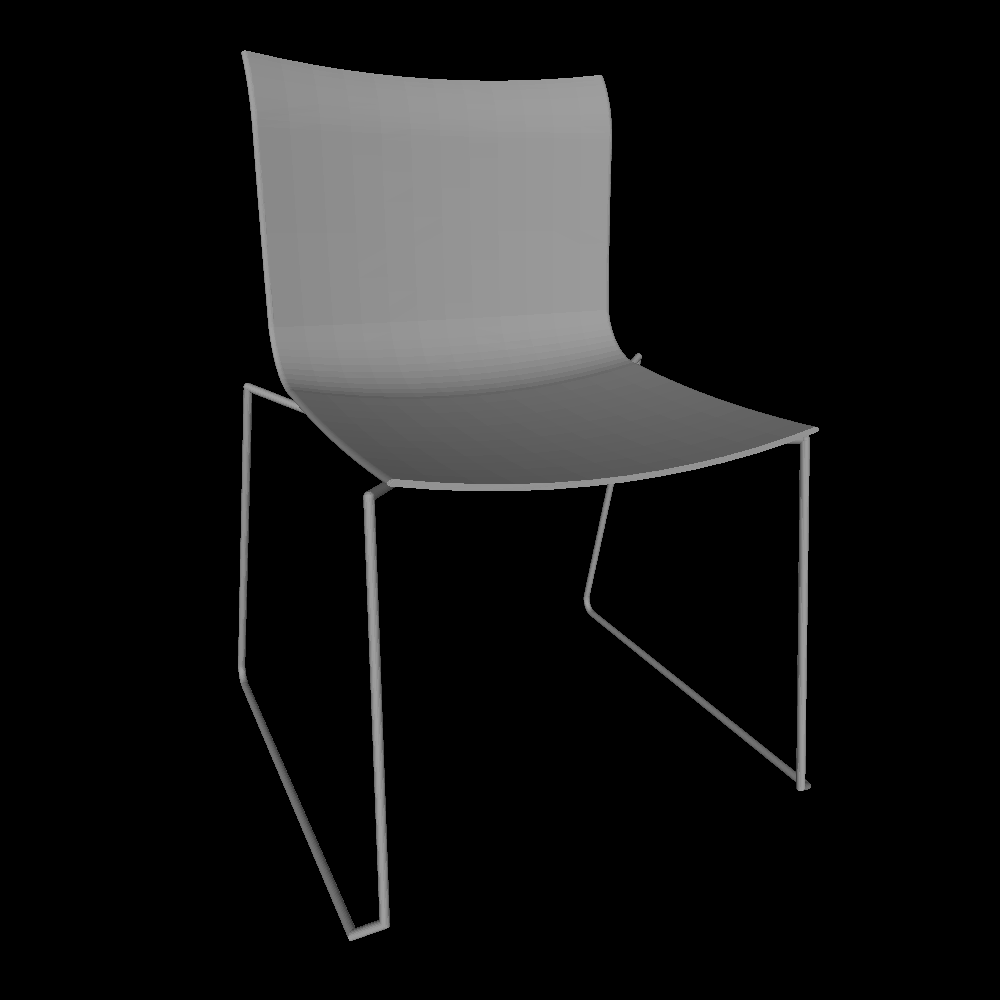}
\end{center}

\textbf{Table example:}

\begin{lstlisting}
Add a tabletop_with_side_frame function that can be used to create a top surface for table objects. It should create a flat table top surface that is framed laterally and horizontally by 4 frame bar parts (on the left, right, front and back sides). The surface and bar parts should be at the same y position, although the height of the surface can vary from the size of each bar frame part. One parameter should control the thickness of each bar frame part. One parameter should control the height of the surface part. This function should always return 5 parts.
\end{lstlisting}

\begin{center}
\includegraphics[width=0.32\columnwidth]{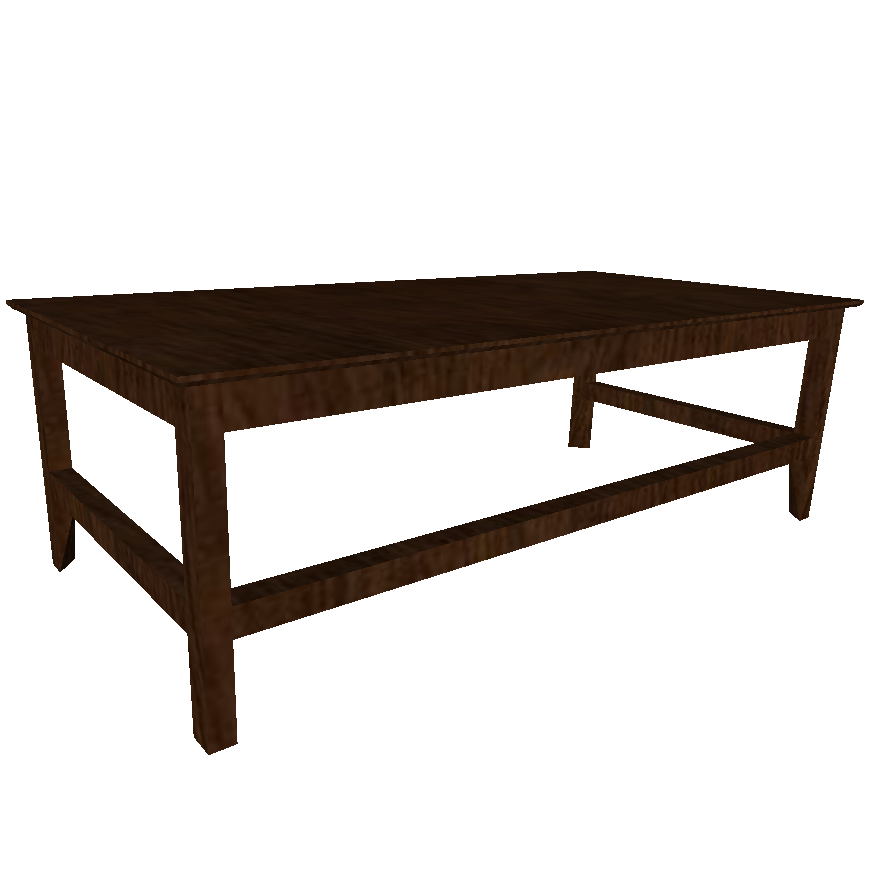}
\hfill
\includegraphics[width=0.32\columnwidth]{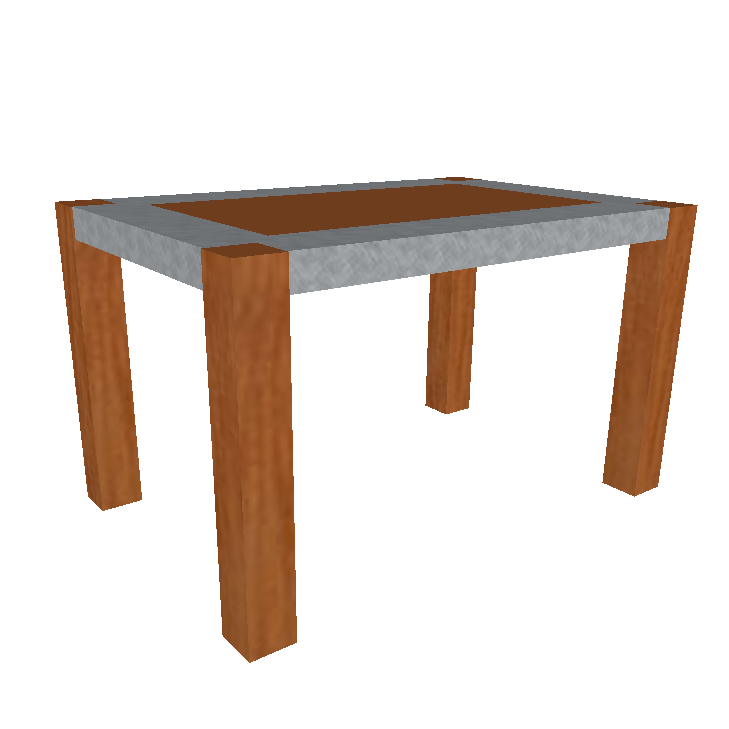}
\hfill
\includegraphics[width=0.32\columnwidth]{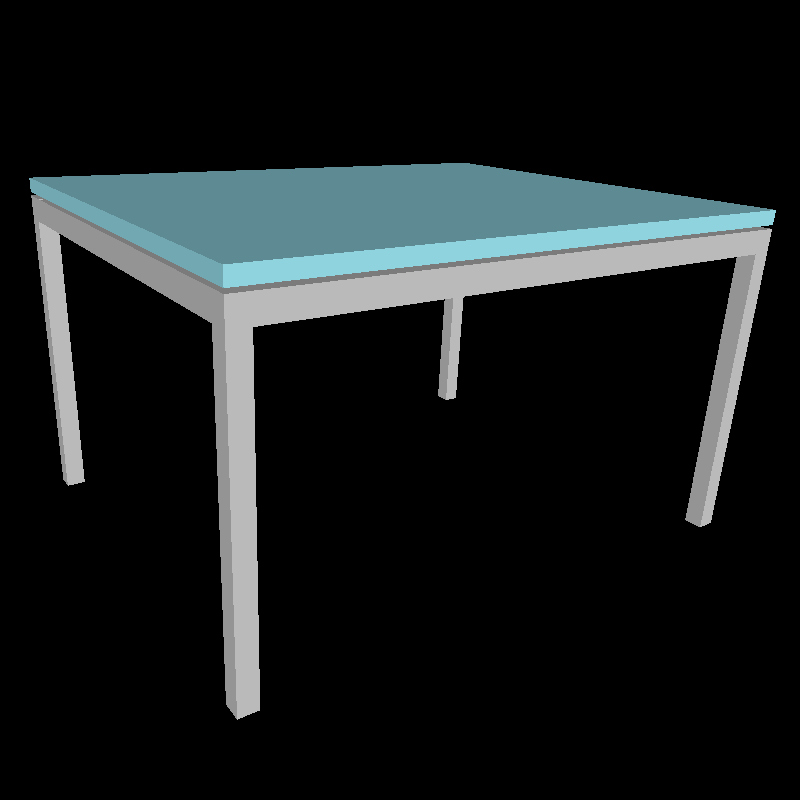}
\end{center}

\textbf{Storage example:}
\begin{lstlisting}
Add a cabinet_drawers function that can be used to create a cabinet drawers for storage furniture objects. It should create a series of vertically stacked drawers, where each drawer has a centrally located handle. One parameter should control the number of drawer units (min of 1, max of 5). One parameter should control the vertical gap in between each pair of drawer units. One parameter should control the depth of each drawer. One parameter should control the width of each handle. One parameter should control the height of each handle. This function will return 2 * the number of drawer units parts.
\end{lstlisting}

\begin{center}
\includegraphics[width=0.32\columnwidth]{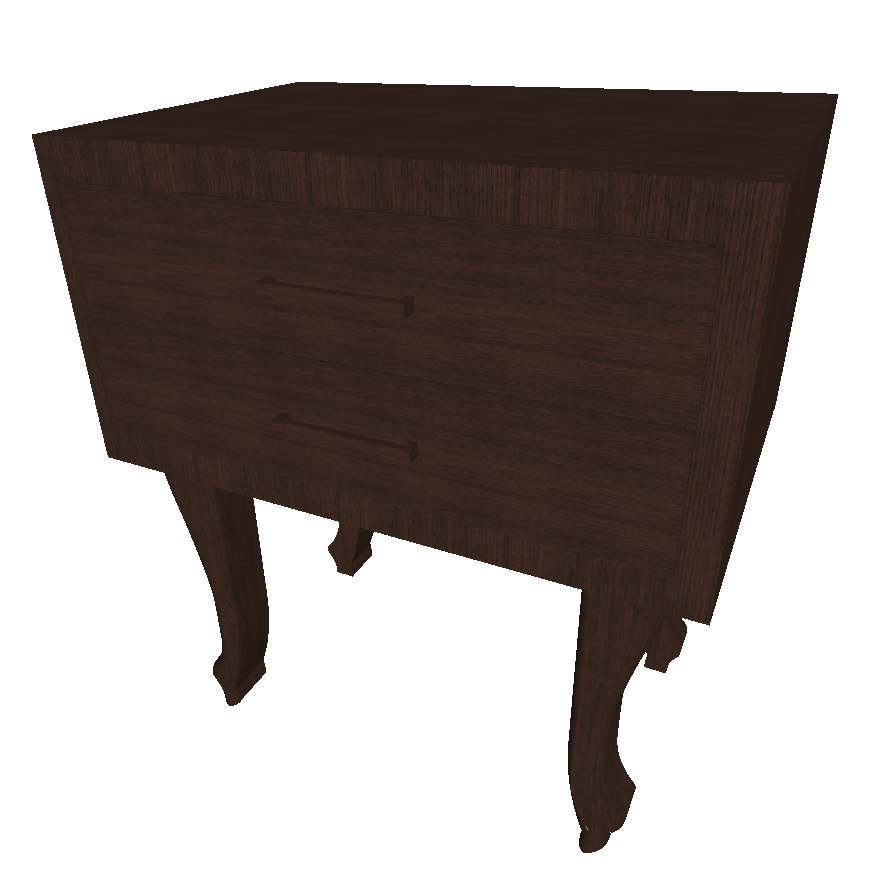}
\hfill
\includegraphics[width=0.32\columnwidth]{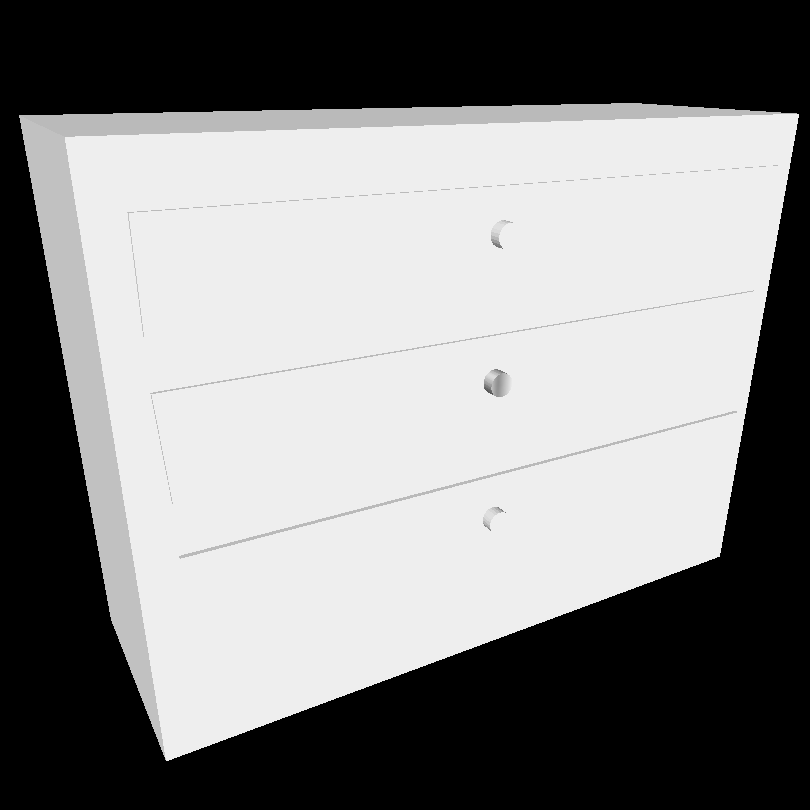}
\hfill
\includegraphics[width=0.32\columnwidth]{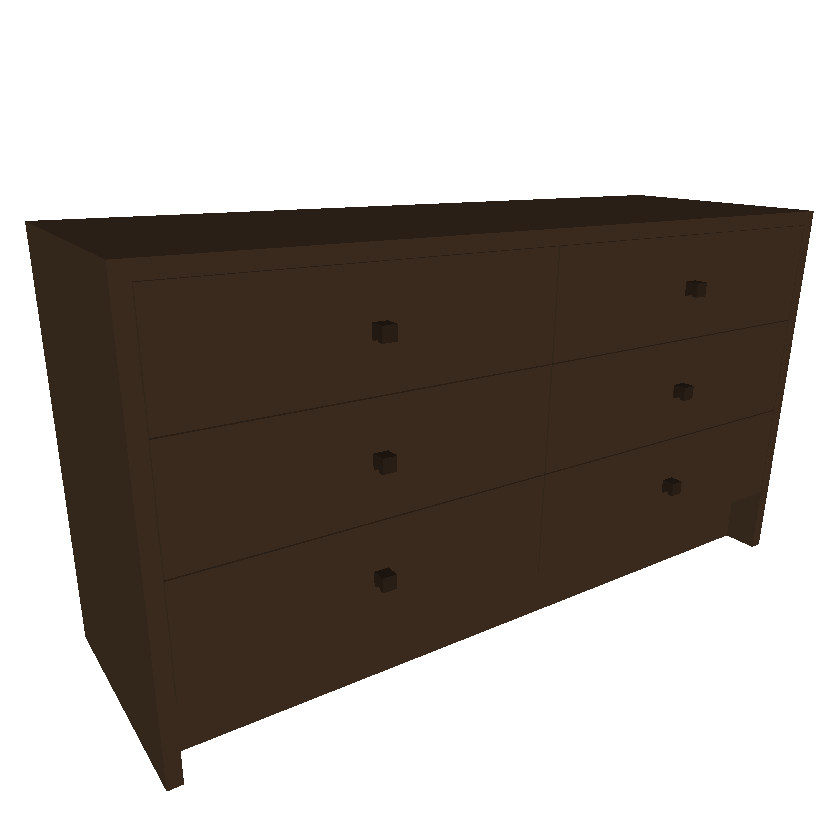}
\end{center}

\textbf{Faucet example:}
\begin{lstlisting}
Add a tube_and_spout function that creates a tube and spout structure for a faucet. The function should consist of two parts: a tube part and a spout part. The width of both parts should span the entire width of the bounding volume. The spout should be positioned at the front of the structure and can either connect to the tube or be contained within it, depending on their relative sizes and positioning. One parameter should control the height of the tube. One parameter should control the depth of the tube. One parameter should control the height of the spout. One parameter should control the depth of the spout. One parameter should control the vertical offset of the spout relative to the tube. The function should always return 2 parts: 1 tube part and 1 spout part.
\end{lstlisting}

\begin{center}
\includegraphics[width=0.32\columnwidth]{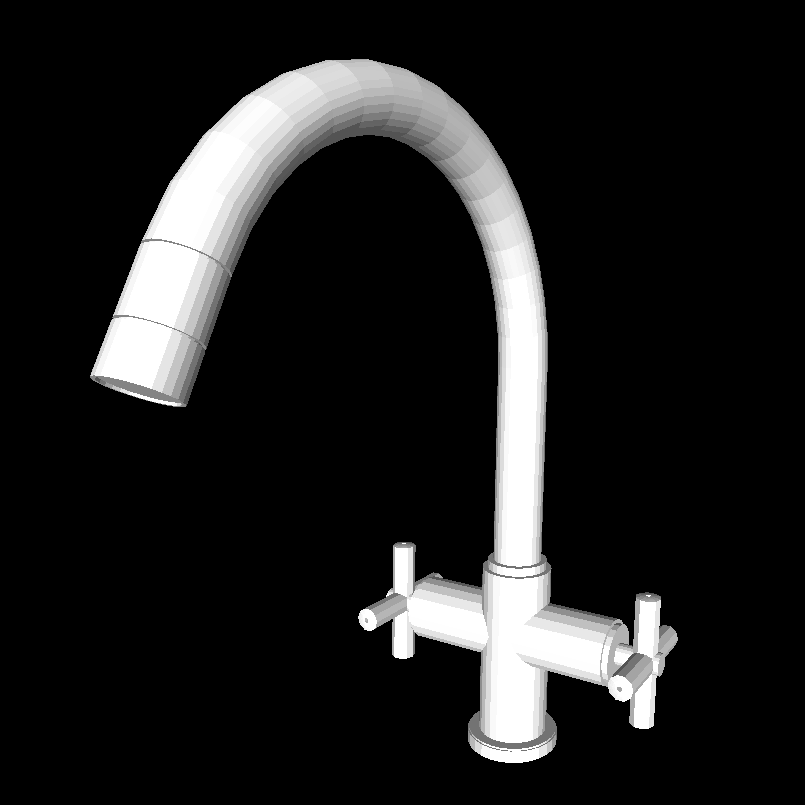}
\hfill
\includegraphics[width=0.32\columnwidth]{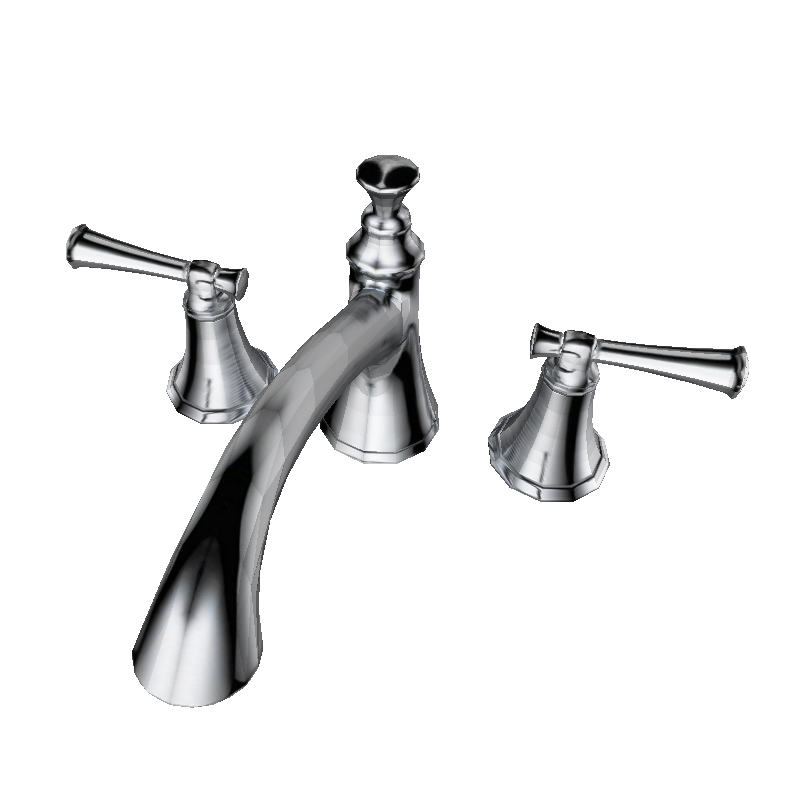}
\hfill
\includegraphics[width=0.32\columnwidth]{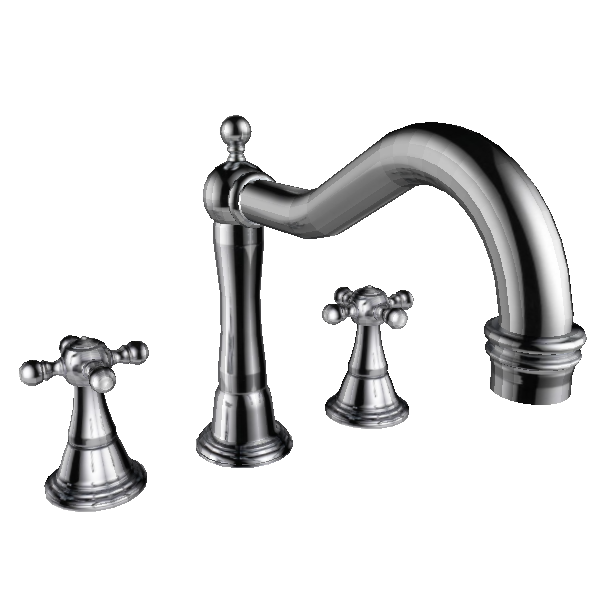}
\end{center}

\textbf{Lamp example:}
\begin{lstlisting}
Add a hanging_lamp function that can be used to create a ceiling-hanging lamp object. It should create three main parts by default: a ceiling mount, a central chain, and a lamp shade. All parts should be vertically aligned and descend from the ceiling in this order. The mount attaches the lamp to the ceiling, with the chain supporting the lamp shade below it. Optionally, a lamp head can be inserted between the chain and the lamp shade. All parts should have square cross-sections (width and depth). The total heights of all included parts should equal the height of the bounding volume. The lamp head is included only if its height and size are both greater than 0.0. One parameter should control the height of the ceiling mount. One parameter should control the size of the ceiling mount. One parameter should control the size of the chain. One parameter should control the height of the shade. One parameter should control the size of the shade. One parameter should control the height of the optional lamp head (default is 0.0). One parameter should control the size of the optional lamp head (default is 0.0). If a lamp head is included, the function should return 4 parts, and if the lamp head is not included, the function should return 3 parts.
\end{lstlisting}

\begin{center}
\includegraphics[width=0.32\columnwidth]{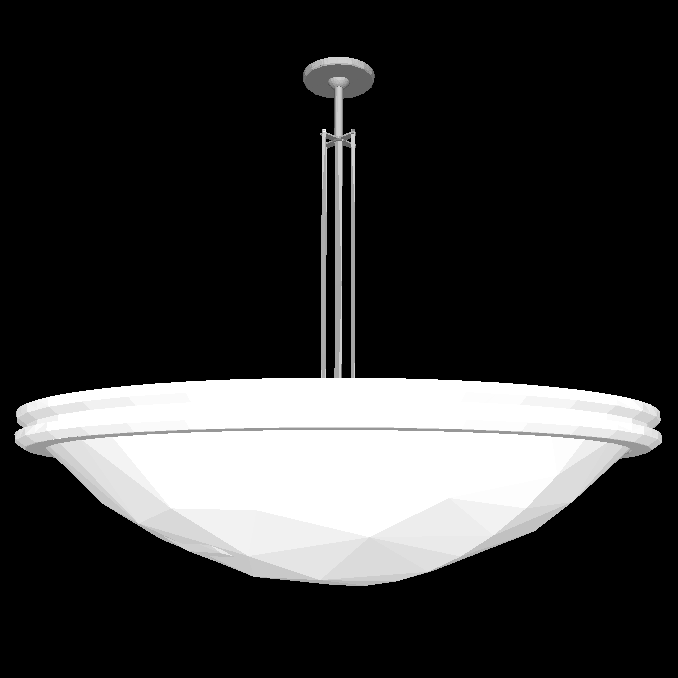}
\hfill
\includegraphics[width=0.32\columnwidth]{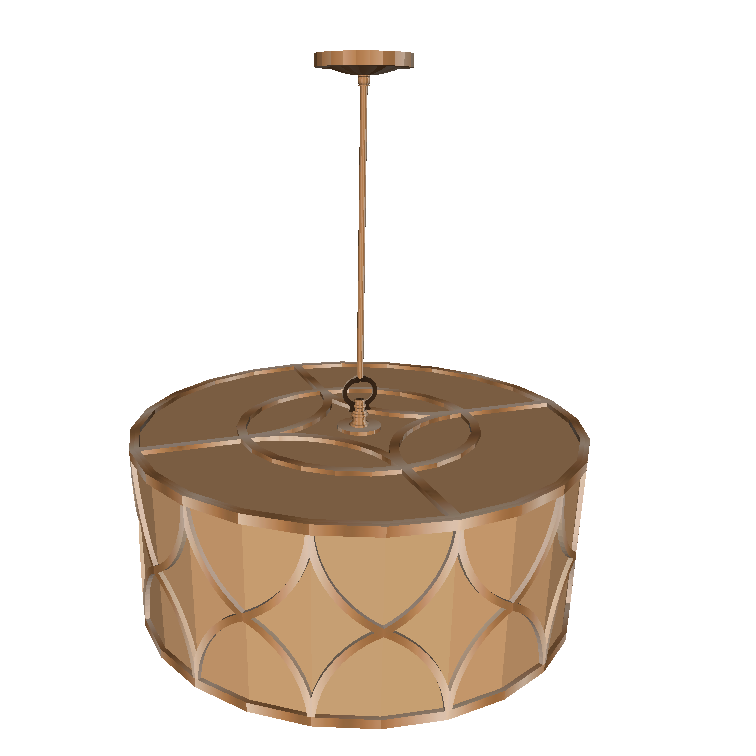}
\hfill
\includegraphics[width=0.32\columnwidth]{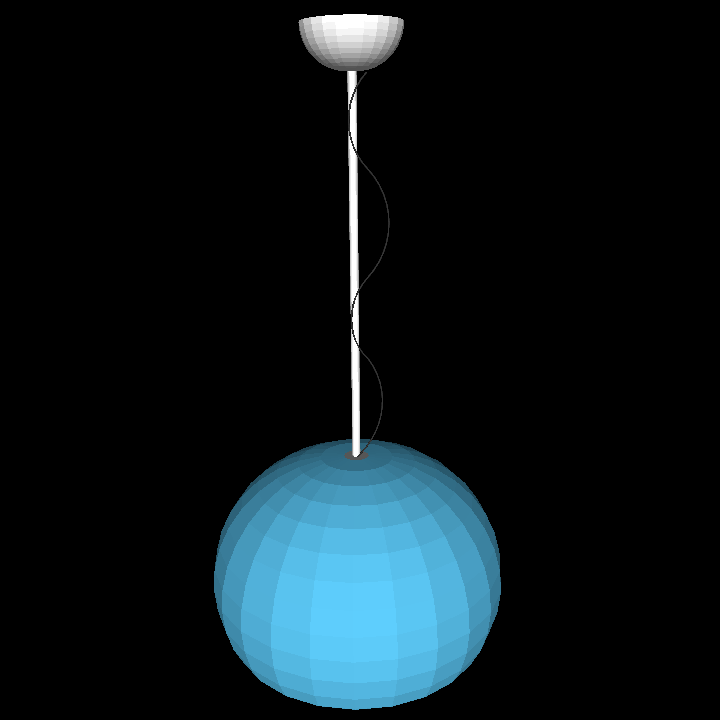}
\end{center}

\end{document}